\documentclass[preprint]{elsarticle}

\usepackage{lineno,hyperref}
\usepackage{amsmath,amssymb} 
\usepackage{caption}

\modulolinenumbers[5]

\journal{Medical Image Analysis}

\biboptions{semicolon,round,sort,authoryear}
\begin{document}

\begin{frontmatter}

\title{MRI Super-Resolution with GAN and 3D Multi-Level DenseNet: Smaller, Faster, and Better}

\author[1,2]{Yuhua Chen}
\ead{chyuhua@ucla.edu}

\author[2]{Anthony G. Christodoulou}

\author[2]{Zhengwei Zhou}

\author[2]{Feng Shi}

\author[2]{Yibin Xie}

\author[1,2]{Debiao Li\corref{corr}}
\ead{debiao.li@cshs.org}

\address[1]{Department of Bioengineering, University of California, Los Angeles, CA, U.S.A}
\address[2]{Biomedical Imaging Research Institute, Cedars-Sinai Medical Center, CA, U.S.A}
\cortext[corr]{Corresponding author}

\begin{abstract}
High-resolution (HR) magnetic resonance imaging (MRI) provides detailed anatomical information that is critical for diagnosis in the clinical application. However, HR MRI typically comes at the cost of long scan time, small spatial coverage, and low signal-to-noise ratio (SNR). Recent studies showed that with a deep convolutional neural network (CNN), HR generic images could be recovered from low-resolution (LR) inputs via single image super-resolution (SISR) approaches. Additionally, previous works have shown that a deep 3D CNN can generate high-quality SR MRIs by using learned image priors. However, 3D CNN with deep structures, have a large number of parameters and are computationally expensive. In this paper, we propose a novel 3D CNN architecture, namely a multi-level densely connected super-resolution network (mDCSRN), which is light-weight, fast and accurate. We also show that with generative adversarial network (GAN)-guided training, the mDCSRN-GAN provides appealing sharp SR images with rich texture details that are highly comparable with the referenced HR images. Our results from experiments on a large public dataset with 1,113 subjects showed that this new architecture outperformed other popular deep learning methods in recovering 4x resolution-downgraded images in both quality and speed.
\end{abstract}


\end{frontmatter}


\section{Introduction}

High spatial resolution MRI provides important structural details for clinicians to detect disease and make better diagnostic decisions~\citep{pruessner2000volumetry}. It provides accurate tissue and organ measurements that benefit quantitative image analysis for better diagnosis and therapeutic monitoring~\citep{park2003super,greenspan2008super,xie2016accounting}. However, limited by hardware capacity and patient cooperation, HR imaging is burdened by long scan time, small spatial coverage, and low signal-to-noise ratio (SNR)~\citep{shi2015lrtv}. HR MRI is also susceptible to respiratory or internal organ motion~\citep{zhou2017optimized,pang2016high}, thus it is very difficult if not impossible to perform on moving part of the body~\citep{stucht2015highest,yang2016free}. In MRI, the duration between phase encodes is the most time-consuming part of the acquisition process, so scan time increases as spatial resolution improves along phase-encoded dimensions. For example, A 4x resolution-degraded LR MRI would be 4x faster than full-resolution HR, at the cost of losing fine local details. Therefore, with the capability to restore resolution loss in HR from just a single LR image, Singe Image Super-Resolution (SISR)~\citep{glasner2009super} is an appealing approach as it promises a reconstructed HR image without adding extra scans or additional multi-image combination processing.

However, the SISR problem is very challenging. Since multiple HR images can be resolution-degraded to the same LR image, SISR is an ill-posed inverse problem. To correctly recover high-frequency details such as local textures and edges from its LR counterparts, an intricate image prior is essential. Previous SISR approaches focus on creating a convex optimization process to find the most likely mapping between LR and HR images~\citep{shi2015lrtv}. Constraints are usually applied to regularize such processes. However, the prior knowledge presumed by those constraints does not always hold. One of the popular regularization methods, total variation~\citep{rudin1992nonlinear}, assumes that the HR image is constant in a small neighborhood, which usually violates the fact that the HR image often carries rich local details and tiny structures, such as intracranial vessels in brain MRI.

In 2D generic images, \citet{dong2016image,dong2016accelerating} show that by utilizing a CNN, the SISR puzzles can be solved with an end-to-end learning-based method. Though a larger neural network with more capacity could help improve the overall performance~\citep{sun2016depth}, training such a deep CNN has been proven to be difficult~\citep{glorot2010understanding}. Recently, with skip connections~\citep{srivastava2015training,he2016deep}, embedding~\citep{szegedy2017inception}, and normalization~\citep{ioffe2015batch}, effective training for a deep neural networks is now made possible. \citet{kim2016accurate} showed that a deeper network using all these advanced techniques could achieve significant improvement in SR image quality, showing that the CNN's architecture is the key to obtain high-quality SR outputs. However, as the network grows deeper, the high-level portion of the network is less likely to make full use of the low-level features due to the vanishing gradient phenomenon~\citep{he2016deep}. Residual learning via skip connection~\citep{ledig2017photo} helps to ease the effect. Later, \citet{huang2017densely} proved that directly stacking all inputs with CNN feature maps strengthens the information flow, and further reduces gradient vanishing. Additionally, these concatenated layers share features more efficiently, lessen the requirement for the immense amount of parameters usually found in deep neural networks. Hence, Densely connected network (DenseNet), can outperform deep CNNs despite its lighter weight. In SISR, \citet{tong2017image} proposed SRDenseNet which combines different hierarchy level features into the final reconstruction layer. Their work demonstrates a significant improvement over networks only using high-level features, indicating that multi-level feature fusion is indeed beneficial for the SISR problem. However, there is still room for SRDenseNet to improve, as we will show in the later section. 

Following the wave of the rapid progress in natural images, SISR has also been adapted into medical image fields~\citep{litjens2017survey,oktay2016multi,you2019ct}. Most of the existing studies directly borrow the 2D network structure and apply it to medical images slice by slice~\citep{oktay2016multi,wang2016accelerating}. However, medical images like Computed Tomography (CT), MRI, and Positron Emission Tomography (PET), often carry anatomy information in 3D. To fully resolve the ill-posed SR problem, a 3D model is more natural and preferable as it can directly extract 3D structural information. Recent studies~\citep{chen2018brain,pham2017brain} show that in brain MRI SR, a 3D CNN outperforms its 2D counterpart by a large margin. However, due to the extra dimension introduced by 3D CNN, the parameter number of a deep model also grows at a staggering rate, the so-called curse of dimensionality. For example, a 3D Fast Super-Resolution CNN (FSRCNN)~\citep{dong2016accelerating} has $5$x parameters than a 2D FSRCNN. Almost all recent SISR methods obtain improved performance by adding more weights and layers~\citep{lim2017enhanced,tai2017image}. However, borrowing such idea to 3D is not ideal. An over-parameterized 3D model is much more heavily weighted, computationally expensive, and less practical with the potential of exceeding the computer's memory limitation. 
Besides, most of the previous CNN SISR approaches are optimized by the pixel/voxel-wise rectilinear or Euclidean distance ($L_1$/$L_2$ loss) between model output and ground truth image. As noticed in \citet{ledig2017photo}, this loss and its derived Peak Signal to Noise Ratio (PSNR) cannot accurately reflect the perceptual quality of the reconstructed image~\citep{johnson2016perceptual}. Therefore, merely taking account of the intensity difference results in suboptimal fuzzy output. 

In this paper, we propose a 3D Multi-Level Densely Connected Super-Resolution Network (mDCSRN) and mDCSRN-GAN with an adversarial loss guided training. Our goal is to build a small, fast, but accurate network structure for the SISR system that can recover 3D details from resolution-reduced MRI. We first experimented with our mDCSRN with $L_1$ loss. Measured by numeric metrics, our mDCSRN outperformed interpolation and popular neural networks while using minimal computational resources. Then we experimented that when trained with a Generative Adversarial Network (GAN)~\citep{goodfellow2014generative}, our mDCSRN-GAN provided even sharper and abundant detailed texture SR images that are highly comparable with the HR images. 

  We summarize four main contributions of this work:
  \begin{itemize}
  \item We proposed a 3D multi-level densely connected super-resolution neural network (mDCSRN) which has  multi-level direct access to all former image features. It is efficient in memory usage yet provides high-quality SR images, making it practical for 3D medical image data.
  \item We proposed a bottleneck compressor module with a fixed-number width before each DenseBlock, which helps balance the layer size in different conceptual levels. The compressor greatly reduces memory usage and increases runtime speed without sacrificing performance.
  \item We proposed a direct combination mechanism that actively feeds all levels' image feature to the final output. This design enables unobstructed gradient flow for easier training and faster convergence. It also makes use of the effect of model ensemble, further boosting performance.
  \item We proposed an mDCSRN-GAN that can produce accurate and realistic-looking SR images by applying a 3D generative adversarial network (GAN) during training. Testing on real-world data showed that our GAN network is robust across different platforms and scanners.
  \end{itemize}

\section{Related Work}

\textbf{Single Image Super-Resolution.} As a classic problem in computer vision, SISR has been studied for decades. Before deep learning approaches dominated the state-of-the-art performance, SISR techniques mostly relied on interpolation, edge-preservation, statistical analysis, and sparse dictionary learning, which have been well-summarized by \citet{yang2014single}. \citet{dong2016image} were the first to propose a SISR based on a three-layer CNN. They showed that a neural network, namely a Super-Resolution Convolutional Neural Network (SRCNN), is naturally capable of handling feature extraction, feature space building, and image reconstruction together through end-to-end training. SRCNN and its recent version Fast SRCNN (FSRCNN) achieved remarkable performance. Their work has inspired many follow-up studies with more advanced network structures~\citep{kim2016accurate,lim2017enhanced,tong2017image,tai2017image}.

\textbf{Efficient Network with Skip Connections.} The performance of the deep learning model keeps improving. However, most of the achievement is built upon the significantly increased model size, wherein the depth of the network becomes a practical issue. As the back-propagated gradients often vanish in the long pathway, it is unlikely to train very deep CNNs. To address this problem, \citet{srivastava2015training} (Highway Network) and \citet{he2016deep} (ResiNet) proposed the bypassing path, or the skip connection, to add the previous layer to the next for smoother information flows. \citet{huang2017densely} discovered that by concatenating previous layers, the network is more efficient and outperforms ResiNet with less number of parameters. As all segments in a DenseNet are directly linked, the gradient can flow unobstructed. Additionally, the dense connections encourage layers to share their features. It dramatically reduces the number of parameters, making the model computational efficient, more robust to new data, and faster to converge.

\begin{figure*}
\centering
\includegraphics[width=\hsize]{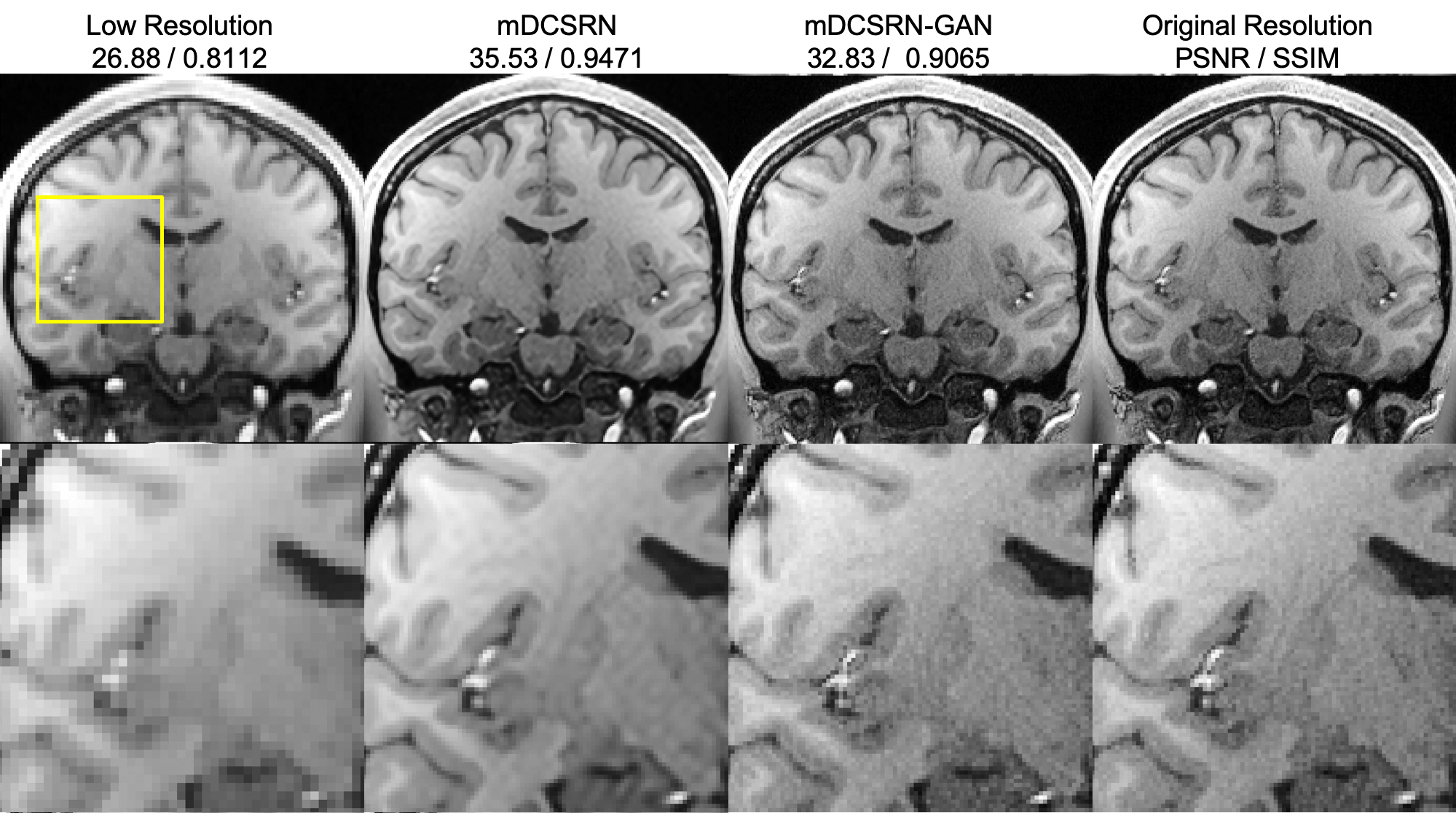}
\caption{Visual quality comparison between Nearest Neighbor Interpolation, deep neural network optimized for intensity difference, deep neural network optimized for a loss with perceptual penalty, and original HR image with PSNR and SSIM shown above the images. (2 x 2 x 1 resolution degrading)}
\label{fig:psnr_demo}
\end{figure*}

\textbf{Super Resolution with Perceptual Loss.} The most straightforward objective for a super-resolution model to optimize would be the voxel-wise difference between model output and the ground-truth image like $L_1$ or $L_2$ loss. However, this difference only takes account of the intensity values' dissimilarity between the reconstructed image and the original image, but not the visual quality which more focuses on sharpness and validity of restored structures. Optimizing the voxel-wise difference will force the model to stack and average all the possible HR candidates in image space. Since a voxel-wise loss doesn't account for the perceptual level of information, despite its results have less intensity error on average, it provides over-blurred and implausible results for the human eye. Therefore, as shown in \textbf{Fig.~\ref{fig:psnr_demo}}, though the voxel-wise loss guided SR model provides a better score in PSNR and SSIM, the model with perceptual loss estimated by a Generative Adversarial Network (GAN) provides more realistic-looking images. 

\section{Method}

We designed our SISR model to learn an accurate inverse mapping of the LR image to the reference HR image during an end-to-end training process. The network is fed with LR images, and it outputs resolution-restored SR images. The HR images were only used in training as the target for the system to optimize. A loss function calculated from SR and HR is back-propagated through layers to adjust weights during training. In the deployment phase, the model only reads LR images and produces SR outputs. We will detail our proposed mDCSRN and the GAN-guided training process in the following sections. 

\subsection{SISR Background}
A SISR system is a feed-forward model to transform an LR image $Y$ into an HR image $X$. A mathematical representation of the resolution downgrading process from $X$ to $Y$ can be written as:
\begin{align}
  Y = f(X),
\end{align}
where $f$ is an arbitrary continuous or discrete function that causes the resolution loss. The SR process is to find an optimal inverse mapping function $g(\cdot)\approx f^{-1}(\cdot)$, where $f^{-1}$ represents the inverse of $f$. The recovered HR image, or SR, $\tilde X$ will be:
\begin{align}
  \tilde X = g(Y) =f^{-1}(Y) + r,
\end{align}
where $r$ is the reconstruction residual. A true inverse $f^{-1}(\cdot)$ does not generally exist, so SISR represents an ill-posed inverse problem.

Despite being ill-posed, the reason why SISR can successfully restore resolution is that both $X$ and $Y$ share information that can be represented in a low-dimensional manifold. A well-trained SISR model should be able to extract visual features from $Y$ and map it into an image feature space. Then $X$ can be reconstructed from the manifold with correct feature mapping. \citet{dong2016image} have shown that CNNs have a built-in nature for the above processes. In a CNN-based SISR technique, all three different steps are trained together:  feature extraction, manifold learning, and image reconstruction. This mingling of different components requires the network to extract the representative low-level feature, construct representative feature space, and precisely reconstruct images from features, which makes CNN based approach achieves state-of-art performance~\citep{dong2016accelerating,kim2016accurate}. 

\subsection{GAN-based Super-Resolution}
\begin{figure}
\centering
\includegraphics[width=0.8\hsize]{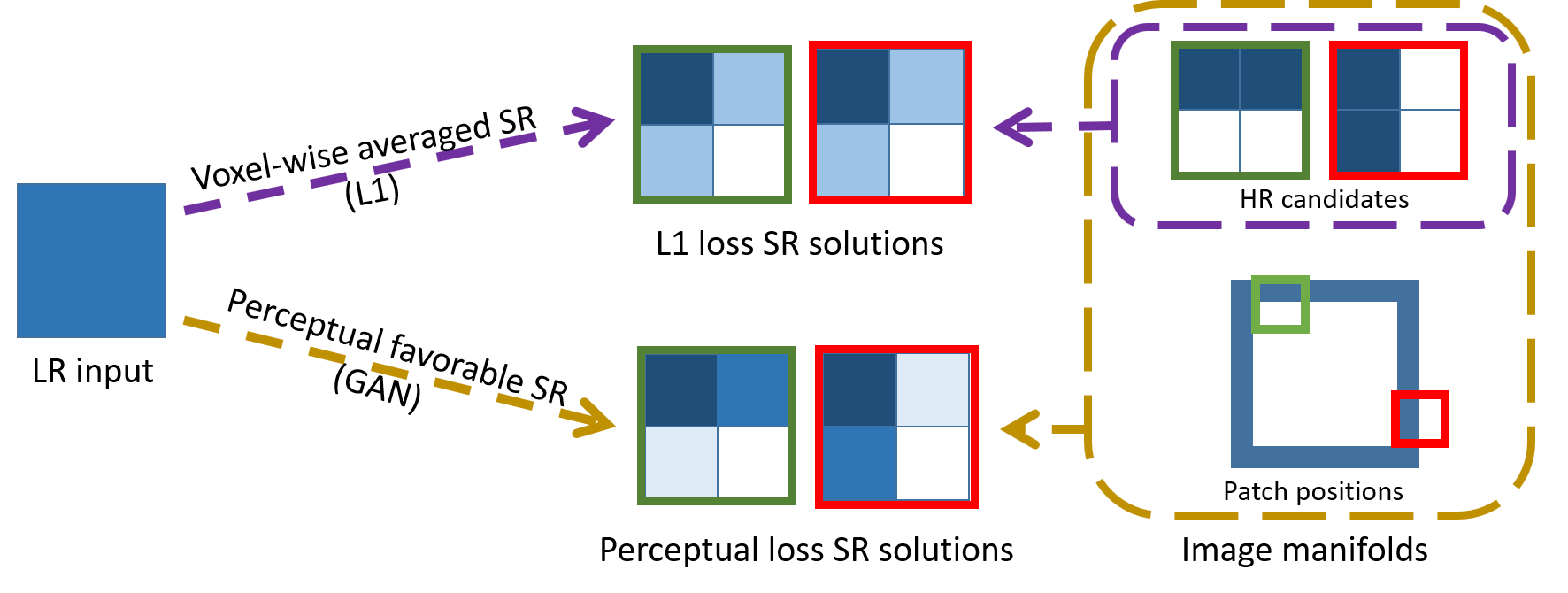}
\caption{An example when an SR model is optimized by $L_1$ vs. perceptual loss in a $2\times2$ neighborhood. The down-sampled LR is the same from two HR patches. Instead of voxel-wisely averaging all possible HR candidates which causes over-smoothing, GAN drives towards perceptual favorable SR solutions by taking account of other informations (i.e. positions) and features in the image manifolds.}
\label{fig:l1_loss_sr}
\end{figure}

Most of the previous SISR approaches optimize the reconstruction by minimizing the voxel-wise difference ($L_1$ or $L_2$ loss) between $\tilde X$ and $X$. However, \citet{ledig2017photo} points out that merely taking care of local voxel-wise differences cast extreme difficulty in restoring important small details due to the ambiguity of the mapping between $X$ and $Y$. We demonstrate one toy example in \textbf{Fig.~\ref{fig:l1_loss_sr}}, where the HR image is $2\times2$ down-sampled to an LR image, and the neighborhood is only in $2\times2$ pixels. When only guided with $L_1$ loss, the SR model doesn't have enough contextual information to recover local neighbors fully. By minimizing the Euclidean loss, it tends to average all possible HR candidates, resulting in a blurred output. However, if we put global perceptual constraints into the account, the SR model is guided by both local intensity information and patch-wise perceptual information, possibly making SR sharper and better-looking. However, such guiding is impossible to be handcrafted because there is no well-adapted mathematical definition of good perceptual quality for images. Based on this observation, \citet{ledig2017photo} proposed to use a Generative Adversarial Network (GAN) for its unsupervised-learning potential of capturing perceptually important image features.

 The GAN framework proposed by \citet{goodfellow2014generative} has two networks: a generator $G$ and a discriminator $D$. The principle of a GAN is to train a $G$ that generates fake images as real as possible, while simultaneously to train a $D$ to distinguish the genuine of them. After training, $D$ becomes very good at separating real and generated images, while the $G$ learns to produce realistic-looking images by the "instruction" from $D$. GAN can model the image representation in an unsupervised manner that doesn't require a pre-designed objective. It is a perfect fit for a SISR. SRGAN~\citep{ledig2017photo} was proposed and shows that the SR model yields unprecedented perceptual quality with the help of GAN.

  However, training a GAN could be very challenging. The balance between $G$ and $D$ has to be carefully maintained so that both of them evolve together. Otherwise, if either side of the lever is too strong, the training quickly landslides to one side, resulting in an under-trained generator $G$ ~\citep{salimans2016improved}. A lot of efforts have been made to stabilize the GAN's training. However, those approaches are highly dependent on the specific network structure, and barely any research has investigated a 3D GAN network. \citet{arjovsky2017wasserstein} observed that the collapse of vanilla GAN training is caused by its optimization toward Kullback-Leibler (KL) divergence between the real and generated probability when there is little or no overlap between them, which is very common at the early stage of training; the gradient from $D$ vanishes, which causes the training to halt. To address this issue, they proposed Wasserstein GAN (WGAN), whose objective is to minimize an efficient approximation of Earth Mover (EM) distance. They proved that this change could remove the difficult-to-achieved requirement for balancing $D$ and $G$. The WGAN enables almost fail-free training in any situation while keeping the quality as good as a vanilla GAN. Additionally, the EM distance from $D$ can also indicate the output image's quality, which is very useful for training.

\subsection{Proposed 3D Multi-Level Densely Connected Super-Resolution Network (mDCSRN)}

Our proposed mDCSRN uses a DenseNet~\citep{huang2017densely} as the starting point. By adding a multi-level densely connection and compressor in each Densely Connected Block (DenseBlock), our network is even more memory-efficient than the original DenseNet and provides excellent images in 3D SISR. An overview of our framework is shown in~\textbf{Fig.~\ref{fig:mDCSRN}}.  All DenseUnits have a growth rate k = 12. We chose exponential linear units (ELU)~\citep{clevert2015fast} as the activation layer to make use of negative values of normalized MRI. We placed a stem module that contains a convolution layer with 2k filters before the feature mapping network, which is a set of densely connected DenseBlocks.  The last part of our mDCSRN is the reconstruction module, which forms the final output. All convolutional layers are using $3\times3\times3$ kernels, except those in the compressor within the DenseBlock and the direct combination layer in the reconstruction module, where kernel size is $1\times1\times1$. There is no up-sampling layer in mDCSRN. As the resolution loss in LR MRI is not in the spatial domain but the k-space, both LR and HR MRI are often generated with the same matrix size when directly fetched from a scanner. We want to discuss structure details as following:

\begin{figure}
\centering
\includegraphics[width=\hsize]{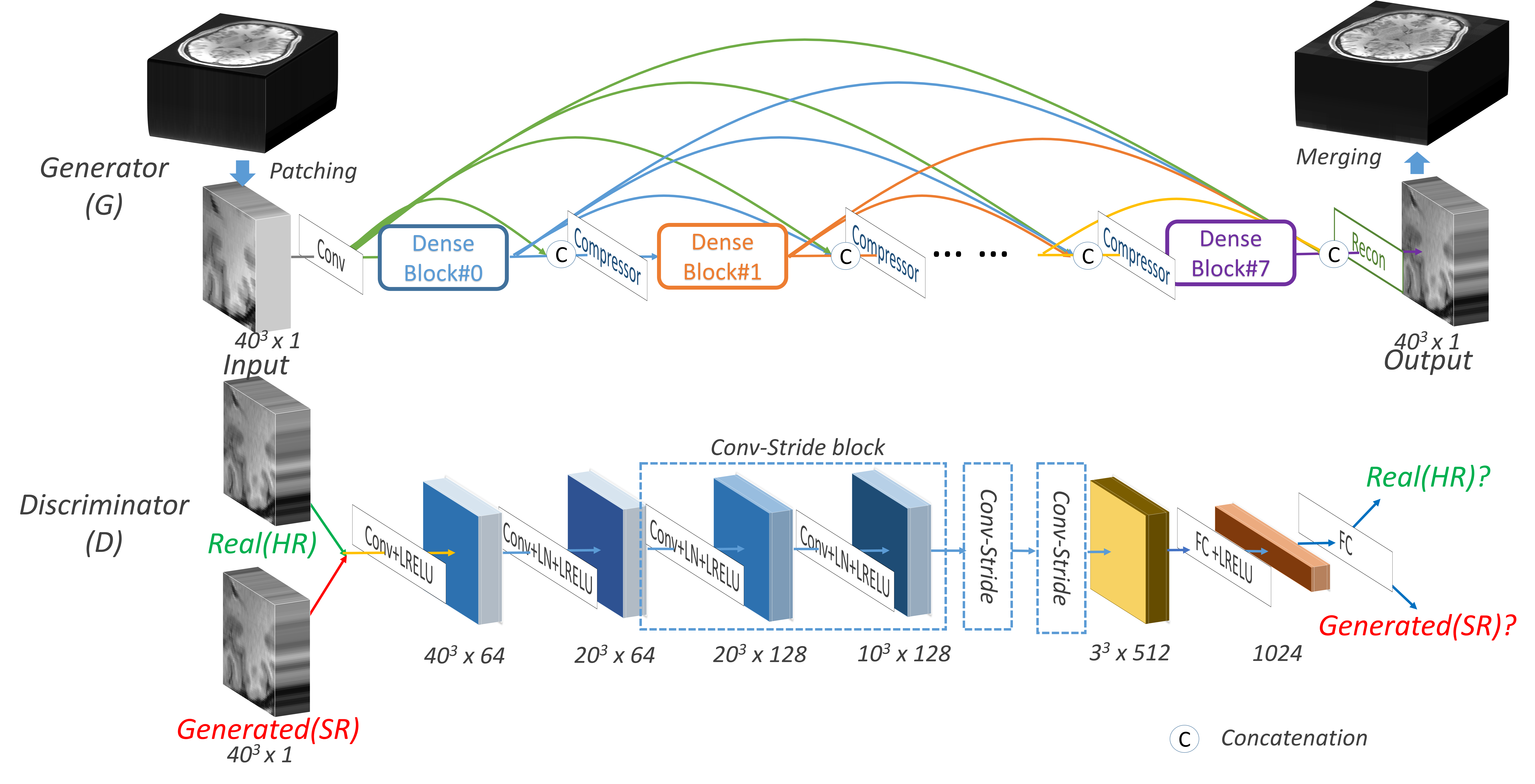}
\caption{mDCSRN-GAN overview. The Generator is our proposed mDCSRN. The Discriminator is adapted from \citet{ledig2017photo}.}
\label{fig:mDCSRN}
\end{figure}

\begin{figure}
\centering
\includegraphics[width=0.6\hsize]{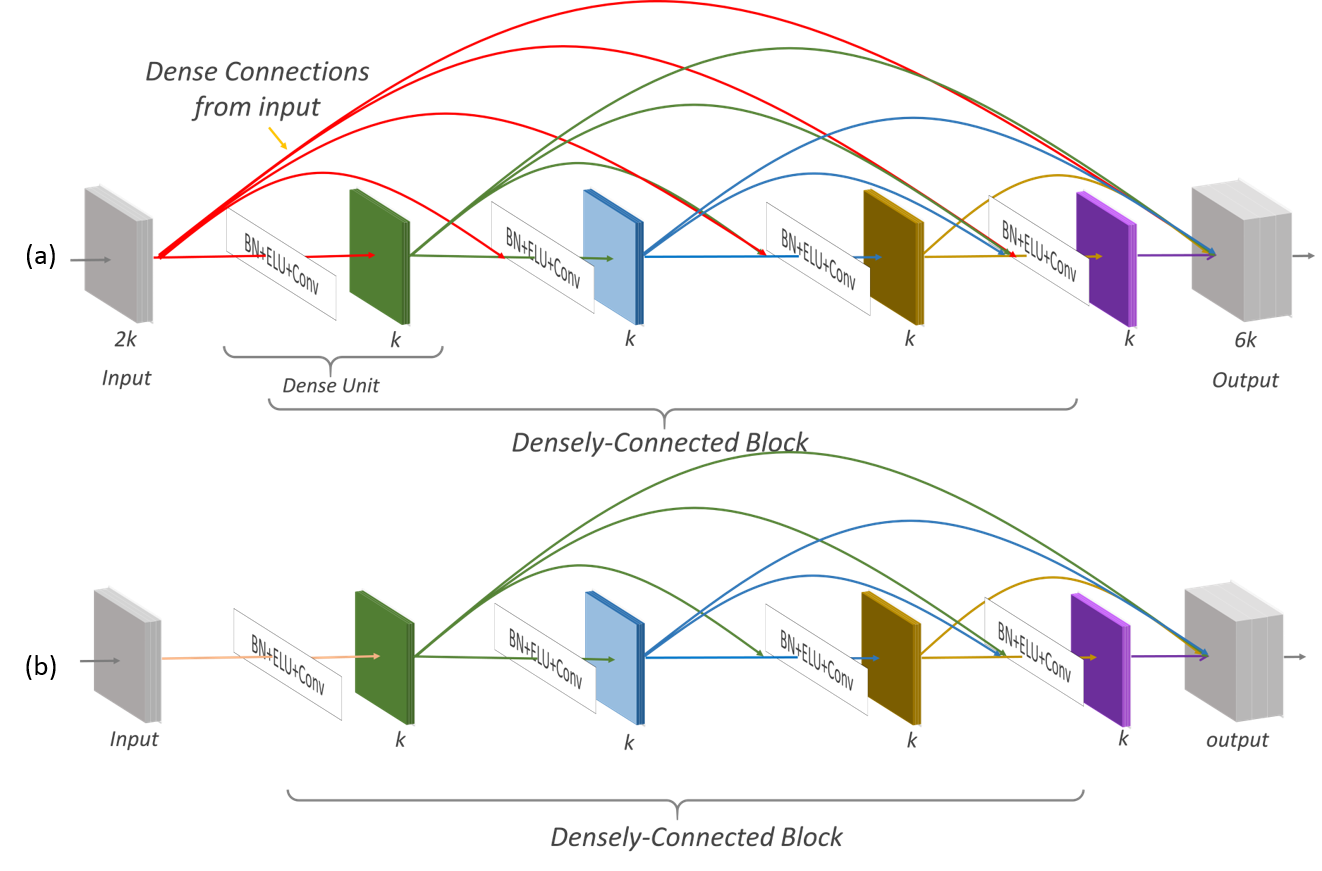}
\caption{Two connectivity ways of a DenseNet: (a)  our proposed mDCSRN vs (b) SRDenseNet~\citep{tong2017image}. Dense connections from the input(red lines in (a)) are missing in (b), which eliminates the direct link to the preceding DenseBlocks.}
\label{fig:dense_block}
\end{figure}

\textbf{Fully Densely Connected Block.} The backbone of the mDCSRN is the DenseBlock from DenseNet~\citep{huang2017densely}. We fully connected all layers within DenseBlocks. It helps to increase feature sharing, making the neural network fewer parameters to keep the same representation capacity. As shown in \textbf{Fig.~\ref{fig:dense_block}}, in our implementation, the input feature map is always directly connected to every convolutional layer, including the output within the DenseBlock, while in \citet{tong2017image} these connections are missing. Those direct links ensure that each DenseUnit can access not only preceding layers within the same DenseBlock but also those in the preceding DenseBlocks, and lead to higher efficiency in parameter usage. To further reduce memory usage, as mentioned in DenseNet-bc~\citep{huang2017densely}, we also put a $1\times1\times1$ bottleneck layer with 4k width before each $3\times3\times3$ convolution when needed. 

\textbf{Multiple Hierarchy Level with Fully Dense Connections.} \citet{veit2016residual} found that Highway Network~\citep{srivastava2015training} and ResiNet~\citep{he2016deep} with skip connections act equally as an ensemble of multiple shallow networks with many paths instead of a giant deep network. Each small network processes some tasks on a different visual level depends on their position. This hierarchical structure harmonizes the animal's visual system discovered by \citet{hubel1962receptive}, which might explain deep ResiNet's excellent performance. As the links within a DenseNet are more effective than ResiNet, this effect is more obvious: all convolutional layer can access all other levels of information and contributes together to the final output. Hence, DenseNet SR is more powerful, as shown in SRDenseNet~\citep{tong2017image}. 

\textbf{Densely Connected DenseBlocks and Compressor.} Though a deep learning model with a single DenseBlock is already capable of providing high-quality SR images~\citep{chen2018brain}, a more sophisticatedly designed architecture still promises better performance. Yet even memory-efficient DenseNets have too many parameters when constructed in 3D. To reduce memory usage while keeping the inter-links strong, we followed the principles of DenseNet and proposed a multi-level densely connected structure. We grouped DenseUnits into DenseBlocks with extra levels of dense connections, as shown in \textbf{Fig.~\ref{fig:mDCSRN}}(G). Then a $1\times1\times1$ convolutional layer (compressor) is applied before each DenseBlock with a fixed output filter number of 2k. According to \citep{szegedy2016rethinking}, this compressor does not negatively affect performance but reduces the weights dramatically. We believe that it brings us at least two advantages: 1) It greatly lessens the parameter number and computation cost; 2) It evens out the weights of different DenseBlocks, forcing the model to focus on low-level features.   

\begin{figure}
\centering
\includegraphics[width=0.6\hsize]{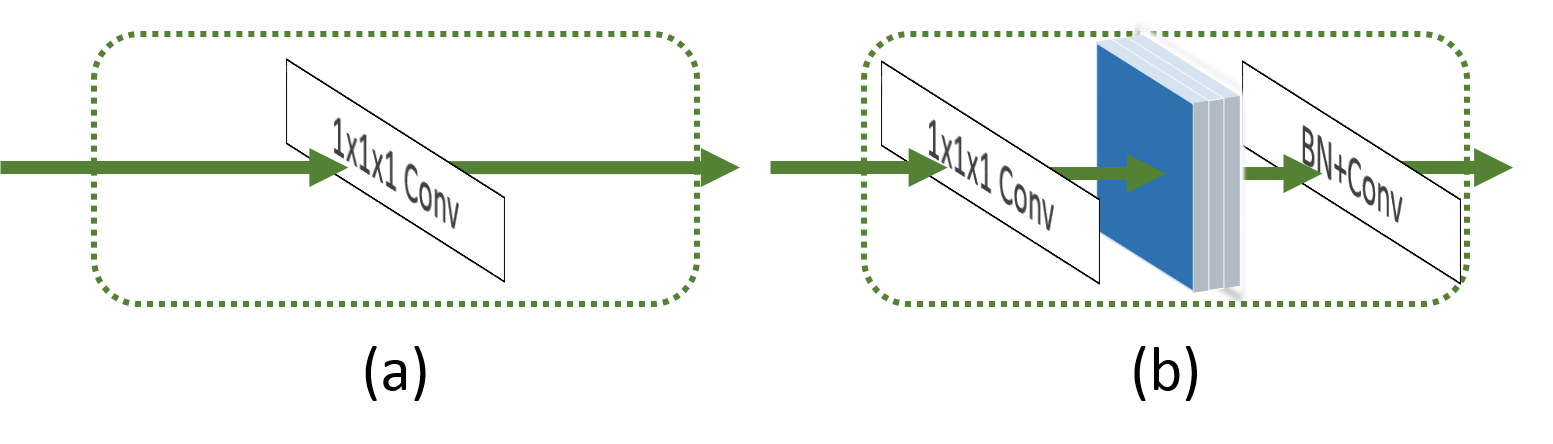}
\caption{Reconstruction network: (a) Directed Feature Combination as proposed in mDCSRN (b) Reconstruction with a bottleneck (8k) followed by a BatchNorm and convolutional layer as proposed in \citet{tong2017image}.}
\label{fig:recon}
\end{figure}

\textbf{Direct Feature Combination.} To further shrink down the model size and improve running speed, in the last module of mDCSRN, we replaced conventional spatial convolutional layers with a 1x1x1 convolutional layer to directly combine all feature maps to the final SR output. This reconstruction process acts as an adaptive feature selection to jointly fuse all the DenseBlock's output. Besides efficiency, as a single DenseBlock is already powerful enough to produce high-quality SR images, our design boosts the ensemble effects of small networks dealing with different visual level information~\citep{liu2016learning}, which conceivably improves SR image quality.  

\textbf{GAN-Guided Training (mDCSRN-GAN).} To achieve plausible-looking SR results, we utilized the adversarial loss from a discriminator in a GAN. The discriminator \textit{D} is built based on the structures of the \textit{D} in SRGAN~\citep{ledig2017photo}. For the type of GAN, we chose WGAN for its excellent stability. Moreover, we use the gradient penalty variant of WGAN, known as WGAN-GP~\citep{gulrajani2017improved}, to accelerate converging in training. As suggested by WGAN-GP, we replace the batch normalization(BN) layer with layer normalization(LN) in the discriminator\textit{D}.

\textbf{Loss Function.}
Our loss function is composed of two parts: intensity loss, $loss_{int}$, and adversarial loss from GAN's discriminator, $loss_{adv}$:
\begin{align}
  loss = loss_{int} + \lambda loss_{adv},
\end{align}
where $\lambda$ is a hyper-parameter, set to 0.1 in experiments. We used the absolute different ($L_1$ loss) between the network output SR and ground-truth HR as the intensity loss:
\begin{align}
  loss_{int} = loss_{L_1} = \frac{\sum_{z=1}^{L} \sum_{y=1}^{H} \sum_{x=1}^{W} | I_{x,y,z}^{HR} - I_{x,y,z}^{SR}|}{LHW},
\end{align}
where $I_{x,y,z}^{SR}$ is the SR and $I_{x,y,z}^{HR}$ is the ground-truth image patch. WGAN's discriminator loss is used as an additional loss in SRGAN network training:
\begin{align}
  loss_{adv} = loss_{WGAN,D} = -D_{WGAN, \theta}(I^{SR}),
\end{align}
where $D_{WGAN, \theta}(I^{SR})$ is WGAN's discriminator output digit for generated SR image patch.

\subsection{LR Image Generation}
An approach to generate LR images from original resolution HR images is required to evaluate the SISR technique. We follow the same steps as in \citet{chen2018brain}: 1) apply 3D FFT to transform HR image into k-space; 2) reduce the resolution by truncating outer part of k-space with a factor of 2x2 in both phase-encoding directions ($2\times2\times1$ ratio in total); 3) convert back to image space by applying inverse FFT and then linearly interpolate to the original image size. This process mimics the actual acquisition of LR and HR images by MRI scanners.

\section{Experiments}
We first describe our experimental settings. Then we conduct a set of experiments to demonstrate that the proposed mDCSRN is not only memory-efficient but also provides state-of-the-art SR results by quantitative metrics. Next, we show that our mDCSRN-GAN provides encouraging qualitative results that are comparable with the ground-truth HR images, as demonstrated by the perceptual scores.

\subsection{Settings}

\textbf{Datasets.} To demonstrate the generalization of mDCSRN, we used the data from the Human Connectome Project (HCP)~\citep{van2013wu}, which is a comprehensive publicly accessible brain MRI database with 1113 subjects. The 0.7 mm isotropic high-resolution 3D T1W images with a matrix size of $320\times320\times256$ were acquired via Siemens 3T Prisma platform on multiple centers. The high-quality ground truth HR images with detailed small structures make this dataset a perfect case to test SISR approaches. The whole dataset is subject-wise split into 780 training, 111 validation, 111 evaluation, and 111 test samples. No subjects nor image patches are overlapped in any subsets. The validation set is used for monitoring and getting the best model checkpoint that has the highest performance during training, measured using mean square error (MSE) for non-GAN training, and EM-distance for GAN training. The evaluation set was used for hyper-parameter searching. The test set is only used for final performance analysis to avoid making model favorable to test data. 

\textbf{Training Details.} The model was implemented in Tensorflow~\citep{abadi2016tensorflow} on a workstation with Nvidia GTX 1080 TI GPUs. For non-GAN networks, ADAM~\citep{kingma1412adam} optimizer with a learning rate of $10^{-4}$ was used to minimize the $L_1$ loss. The batch size was set to 6. We followed a similar process of patching and data augmentation as in \citet{chen2018brain}, except, the patch size during training was set as $40\times40\times40$. We trained mDCSRN for 800k iterations, which is about 300 epochs, as 18 randomly sampled patches were fetched from a patient during training, lasting from 5 to 14 days depending on network size. For GAN experiments, we transfer the weights from well-trained mDCSRN above as the initial $G$ of mDCSRN-GAN. We first trained $D$ for the initial 10k steps without updating $G$. After then, for 5 iterations of training the $D$, $G$ was trained once. Additionally, after every 500 iterations of $G$ training, $D$ was trained for an extra 200 steps. It is solely to make sure $D$ is always ahead of $G$, as suggested in WGAN~\citep{arjovsky2017wasserstein}. Adam optimizer with $5\times10^{-6}$ was used to optimize $G$ for a total of 200k steps. 

\textbf{SR Generation.} Once training was finished, LR images from the evaluation/test set were fed into the model to generate SR outputs. A patch size of $70\times70\times70$ with a margin 3 was used in testing to avoid artifacts on the edges. The merging of the output patches was done without averaging. Because the batch size is 1 during testing, we set the batch normalization layers in the model to "train" mode instead of "test" mode for better estimation. We recorded the runtime speed on a single Nvidia GTX 1080 TI GPU.

\textbf{Quality Metrics.} To quantitatively measure mDCSRN's recovery accuracy, we used three reference-based image similarity metrics: structural similarity index (SSIM)~\citep{wang2004image}, peak signal to noise ratio (PSNR), and normalized root mean squared error (NRMSE). Numbers were calculated in the most resolution degraded cross-section ($2\times2$) slice by slice. Scores were reported in its subject-wise slice-averaged numbers. For mDCSRN-GAN measurement, we list its numeric metrics as well. But we need to point out that PSNR could not fully represent the visual quality. Hence, we measured the perceptual quality via non-reference metrics: PIQE~\citep{venkatanath2015blind}, Ma's score~\citep{ma2017learning}, NIQE~\citep{mittal2012making}, and perceptual index (PI, used in PRIM-SR Challenge~\citep{blau20182018}). To efficiently calculate the perceptual scores, we only processed the 2D slices where the foreground (brain region) occupies more than $25\%$ of the whole image. All perceptual scores were calculated in MATLAB R2019 software. 

\textbf{Segmentation Evaluation.} In the testing stage, to further exemplify the benefits from our SR for the automatic medical image processing system, we conducted a fully automated segmentation on 159 brain tissues from a pre-trained high-performance neural network: HighRes3D~\citep{li2017compactness}. We performed the test on the output of bicubic interpolation, SRResNet, mDCSRN $b8u4$, and mDCSRN-GAN $b8u4$. We first interpolated all images from the original $0.7 mm^3$ spatial resolution into $1.0 mm^3$ since the HighRes3D network was trained on the latter resolution. Then, we performed an N4 bias correction~\citep{tustison2010n4itk} with ANTS~\citep{avants2009advanced} toolbox. Then we ran the inferences of HighRes3D on the NiftyNet~\citep{gibson2018niftynet} open-platform. We used two similarity metrics, Dice Similarity Coefficient (DSC)~\citep{sorensen1948method} and Jaccard Index (JACC)~\citep{jaccard1901etude}, to quantitatively measure the agreement of segmentation between the up-sampled/super-resolution and the high-resolution images. Numbers were average among those 159 different anatomical structures.

\subsection{Results}
We first demonstrate that the compressor in our multi-level densely connection does improve memory efficiency. We show that by replacing spatial convolutional layers with a single direct feature combination, we further reduce the model size without sacrificing performance. We show how the depth and width of mDCSRN affect performance, and we compare mDCSRN with other popular SISR models. Qualitatively, we show the results from the mDCSRN-GAN side by side with other up-sampling methods. The mDCSRN-GAN provides realistic-looking images while running at the same time as our mDCSRN. We further investigate the perceptual quality with quantitative non-reference metrics. To demonstrate our model's clinical value in automatic systems, we use the brain tissue segmentation as an example to demonstrate the benefits brought by SR models. Last, we show that in the real-world scan, our mDCSRN-GAN exhibits its fantastic stability across different platforms.

\begin{table*}
\footnotesize
\begin{center}
\caption{Ablation experiment results of mDCSRN on the \textbf{evaluation} set}
\label{table:evaluation}
\begin{tabular}{llllll}
\hline\noalign{\smallskip}
  & PSNR\ddag & SSIM\ddag & NRMSE\dag & \#Parm & time(s)\\
\noalign{\smallskip}
\hline
\noalign{\smallskip}
\hline
Exp. 1 &  \multicolumn{5}{c} {k=12}  \\
\hline
\textbf{b1u16-r} & $\textbf{35.84}\pm\textbf{0.86}$ & $0.9408\pm0.0060$ & $\textbf{0.0866}\pm\textbf{0.0036}$ & 0.35M & 18.69\\
\textbf{b4u4-r} & $\textbf{35.84}\pm\textbf{0.86}$  & $\textbf{0.9410}\pm\textbf{0.0060}$ & $\textbf{0.0866}\pm\textbf{0.0037}$  & 0.26M & \textbf{12.93} \\
\textbf{b1u12-r} & $35.76\pm0.86$  & $0.9396\pm0.0060$ & $0.0874\pm0.0036$  & 0.25M & \textbf{12.74} \\
\hline
\noalign{\smallskip}
\hline
Exp. 2 & \multicolumn{5}{c} {k=12}    \\
\hline
\textbf{b4u4-r} & $35.84\pm0.86$  & $\textbf{0.9410}\pm\textbf{0.0060}$ & $0.0866\pm0.0037$  & 0.26M & 12.93 \\
\textbf{b4u4} & $\textbf{35.88}\pm\textbf{0.85}$ & $0.9408\pm0.0060$ & $\textbf{0.0864}\pm\textbf{0.0037}$ & \textbf{0.22M} & \textbf{12.06}\\
\hline
\noalign{\smallskip}
\hline
Exp. 3 & \multicolumn{5}{c} {k=12}      \\
\hline
\textbf{b4u4}   & $35.88\pm0.85$ & $0.9408\pm0.0060$ & $0.0864\pm0.0037$ & 0.22M & \textbf{12.06}\\
\textbf{b6u4}   & $36.06\pm0.86$ & $0.9431\pm0.0059$ & $0.0845\pm0.0037$ & 0.35M & 19.14  \\
\textbf{b8u4}   & $\textbf{36.14}\pm\textbf{0.87}$ & $\textbf{0.9442}\pm\textbf{0.0059}$ & $\textbf{0.0836}\pm\textbf{0.0037}$ & 0.49M & 28.52 \\
\hline
\noalign{\smallskip}
\hline
Exp. 4& \multicolumn{5}{c} {b4u4}      \\
\hline
\textbf{k=8}    & $35.57\pm0.85$ & $0.9382\pm0.0060$ & $0.0894\pm0.0036$ & \textbf{0.10M} & \textbf{7.95}  \\
\textbf{k=12}   & $35.88\pm0.85$ & $0.9408\pm0.0060$ & $0.0864\pm0.0037$ & 0.22M & 12.06\\
\textbf{k=16}   & $\textbf{35.96}\pm\textbf{0.87}$ & $\textbf{0.9424}\pm\textbf{0.0059}$ & $\textbf{0.0854}\pm\textbf{0.0037}$ & 0.41M & 15.37  \\
\hline
\noalign{\smallskip}
\hline
Exp. 5& \multicolumn{5}{c} {}      \\
\hline
\textbf{b4u4k12}  & $\textbf{35.88}\pm\textbf{0.85}$ & $0.9408\pm0.0060$ & $0.0864\pm0.0037$ & 0.22M & \textbf{12.06}\\
\textbf{b8u4k8}   & $35.85\pm0.86$ & $\textbf{0.9415}\pm\textbf{0.0059}$ & $\textbf{0.0863}\pm\textbf{0.0037}$ & 0.22M & 18.43   \\
\hline
\textbf{b4u4k16}  & $35.96\pm0.87$ & $0.9424\pm0.0059$ & $0.0854\pm0.0037$ & 0.41M & \textbf{15.37}  \\
\textbf{b8u4k12}  & $\textbf{36.14}\pm\textbf{0.87}$ & $\textbf{0.9442}\pm\textbf{0.0059}$ & $\textbf{0.0836}\pm\textbf{0.0037}$ & 0.49M & 28.52 \\
\hline
\end{tabular}
\end{center}
 {\ddag: The higher the better, ~\dag: The lower the better\\ \raggedright \textbf{b}:\# DenseBlock, \textbf{u}:\# DenseUnit per Block, \textbf{k}: Growth rate \textbf{-r}: with reconstruction layer; default is using direct combination layer \par}
\end{table*}

\textbf{Multi-Level Connectivity and Compressor.} As shown in \textbf{Table~\ref{table:evaluation}} Exp. 1, with the same total number of DenseUnit, mDCSRN b4u4-r had fewer parameters, ran faster, and achieved the same performance as the original DenseNet design b1u16-r; with the same amount of parameters, b4u4-r significantly outperformed b1u12-r; proving that multi-level connectivity and compressor together helped improve memory efficiency and runtime speed.

\textbf{Direct Feature Combination vs Extra Reconstruction Layer.} As shown in \textbf{Table~\ref{table:evaluation}} Exp. 2, with the same depth, b4u4 with our introduced direct feature combination achieved similar to slightly better performance than b4u4-r with reconstruction layers while decreasing model size by 15\%.

\textbf{Depth vs Width.} The results with different depth and width configuration are shown in \textbf{Table~\ref{table:evaluation}} Exp. 3 and Exp. 4. The performance was improved by either making the network deeper or wider, at the cost of more extensive memory consumption and slower inference speed. As shown in \textbf{Table~\ref{table:evaluation}} Exp. 5, when models are in a similar size, the deeper network, the better the performance. Although the weight-saving mechanism is more effective in the deep and narrow network, it runs slower, due to the extra computational cost from additional bottleneck layers. Therefore, given a fixed memory constraint, a shallow mDCSRN is preferable for a fast application, while a deep mDCSRN is excellent for better results.

\begin{table*}
\footnotesize
\begin{center}
\caption{mDCSRN vs. interpolation and previous CNN based SISRs on the \textbf{test} set}
\label{table:test}
\begin{tabular}{llllll}
\hline\noalign{\smallskip}
	\multicolumn{6}{c}{Intensity-based similarity metrics}\\
\noalign{\smallskip}	
\hline
  & PSNR\ddag & SSIM\ddag & NRMSE\dag & \#parm & time(s)\\
\hline
\textbf{NN} & $29.48\pm0.81$ & $0.8219\pm0.0113$ & $0.2007\pm0.0071$ & N/A & N/A\\
\textbf{Bicubic} & $30.30\pm0.82$ & $0.8420\pm0.0105$ & $0.1830\pm0.0067$ & N/A & N/A\\
\hline
\textbf{FSRCNN}   & $34.33\pm0.81$ & $0.9207\pm0.0062$ & $0.1142\pm0.0050$ & \textbf{0.06M} & 15.57\\
\textbf{SRResNet} & $36.09\pm0.82$  & $0.9425\pm0.0052$ & $ 0.0939\pm0.0043$  & 2.01M & 107.16 \\
\textbf{SRDenseNet} & $35.93\pm0.82$  & $0.9413\pm0.0052$ & $0.0955\pm0.0044$  & 0.39M & 17.95 \\
\hline
\textbf{b4u4k12} & $36.08\pm0.82$ & $0.9418\pm0.0052$ & $0.0935\pm0.0044$ & 0.23M & \textbf{12.54}\\
\textbf{b6u4k12} & $36.31\pm0.82$  & $0.9438\pm0.0051$ & $0.0915\pm0.0043$  & 0.35M & 19.60 \\
\textbf{b8u4k12} & $\textbf{36.39}\pm\textbf{0.82}$  & $\textbf{0.9448}\pm\textbf{0.0050}$ & $\textbf{0.0906}\pm\textbf{0.0043}$  & 0.49M & 27.86 \\
\hline
\end{tabular}
\begin{tabular}{lllll}
\hline\noalign{\smallskip}
	\multicolumn{5}{c}{Perceptual quality metrics}\\
\noalign{\smallskip}
\hline
  & PIQE\dag & NIQE\dag & MA's Score\ddag & PI\dag\\
\hline
\noalign{\smallskip}
\textbf{Bicubic} & $99.54\pm1.39$ & $5.53\pm0.13$ & $3.20\pm0.06$ & $6.17\pm0.08$\\
\textbf{SRResNet} & $80.85\pm7.51$  & $5.92\pm0.19$ & $\textbf{5.06}\pm\textbf{0.03}$  & $5.43\pm0.08$ \\
\textbf{mDCSRN $b8u4$} & $81.04\pm7.79$  & $6.01\pm0.19$ & $\textbf{5.06}\pm\textbf{0.03}$  & $5.48\pm0.09$ \\
\textbf{mDCSRN-GAN $b8u4$} & $\textbf{71.86}\pm\textbf{7.24}$  & $\textbf{5.00}\pm\textbf{0.16}$ & $5.04\pm0.04$  & $\textbf{4.98}\pm\textbf{0.07}$ \\
\textbf{Original Resolution} & $\textbf{69.70}\pm\textbf{7.02}$ & $5.52\pm0.14$ & $5.05\pm0.03$ & $5.23\pm0.06$\\
\hline

\end{tabular}
\end{center}
{\ddag: The higher the better, ~\dag: The lower the better \par}
\end{table*}

\begin{figure*}
\centering
\includegraphics[width=\textwidth]{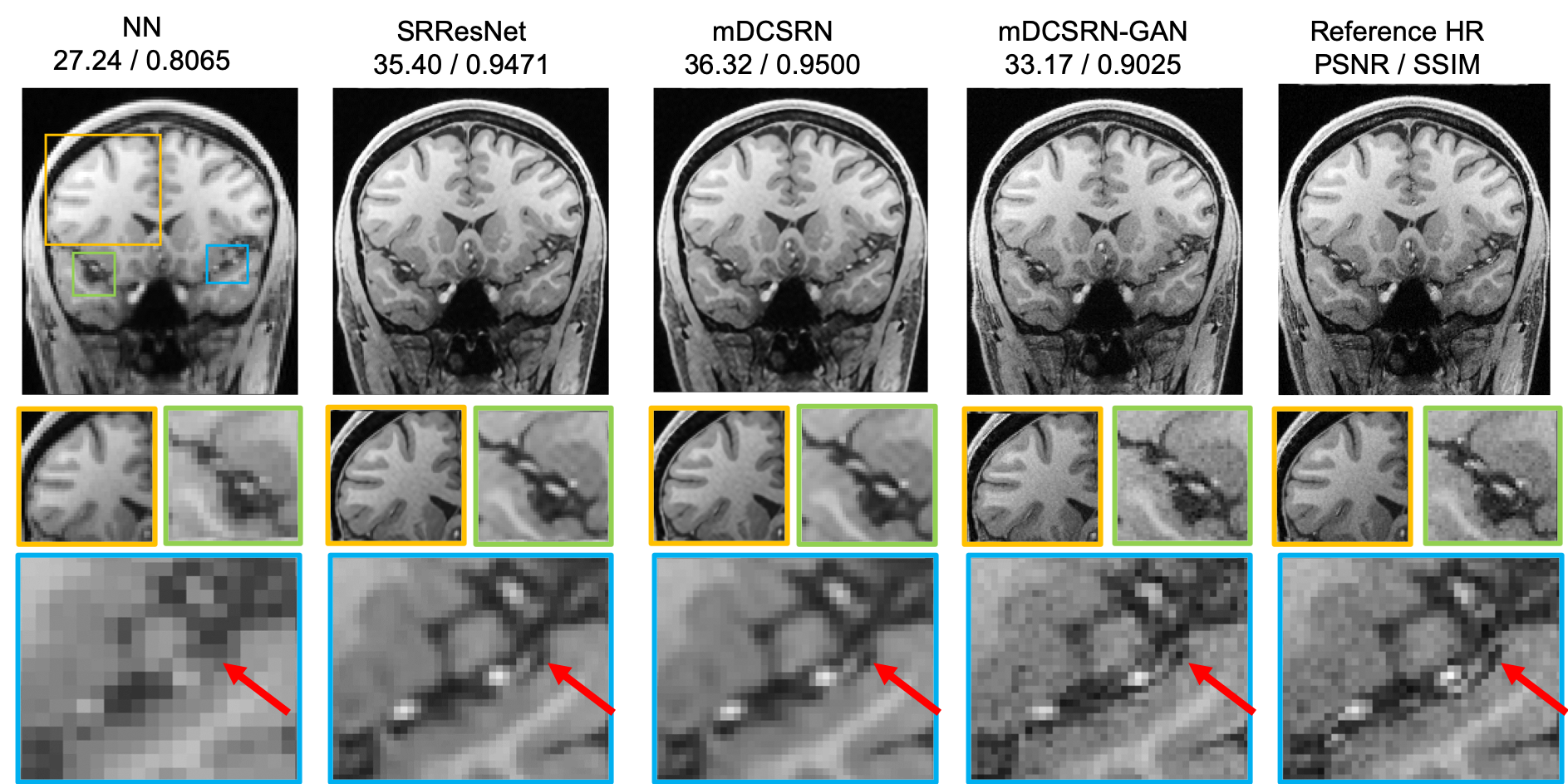}
\caption{Example results from the test set of Nearest Neighbor, SRResNet, mDCSRN $b8u4$, mDCSRN-GAN $b8u4$ in the $2\times2$ resolution degraded plane. PSNR and SSIM of this subject are shown on the top. Despite performing worse in PSNR and SSIM, GAN SR images appear to have recovered more spatial details.}
\label{fig:examples}
\end{figure*}

\textbf{Baseline.} As baseline models, FSRCNN~\citep{dong2016accelerating}, SRResNet~\citep{ledig2017photo}, and SRDenseNet~\citep{tong2017image} were implemented and extended to 3D. As there is no image-size changing in our SISR, the up-sampling CNNs (transposed-convolutional layers or sub-pixel layers) in those original designs were replaced with the same scale convolutional layers. For SRDenseNet, we adjusted the hyperparameters as similiar as possible to mDCSRN $b8u4$ (i.e. reduced DenseUnit number from 8 to 4, changed activation function to ELU, and set growth-rate k=12). All models were trained for 300 epochs. With respect to quantitative similarity metrics, as shown in \textbf{Table~\ref{table:test}}, the lightest mDCSRN $b4u4$ ran fastest among all CNN approaches with competitive results. The deepest mDCSRN $b8u4$ as shown in \textbf{Fig.~\ref{fig:examples}} outperformed all previous SISR approaches by a considerable margin. It did run slower than SRDenseNet but was still 4x faster than the SRResNet. Both SRDenseNet and mDCSRN $b6u4$ are similar in model size and running speed, but the later significantly outperformed the former, proving the advantage of our efficient architecture design.

\textbf{Perceptual Quality.} An example output is shown in \textbf{Fig.~\ref{fig:examples}}. mDCSRN $b8u4$ provides slightly better SR reconstruction accuracy than SRResNet, but it is mDCSRN-GAN $b8u4$ that more closely shapes the small vessel pointed by the red arrows. Though mDCSRN-GAN's PSNR is lower than its non-GAN sibling, it provides more structural details that are more plausible by the human eye. As shown in \textbf{Table~\ref{table:test}}, the quantitative perceptual quality numbers suggest that while non-GAN SR shows slightly closer to HR only in the MA's metric, the GAN SR model shows much better performance in all other three measurements. GAN even obtained a higher score in NIQE and PI than HR, since SR images were generated from less noisy LR input, making the SR more plausible for noise-sensitive perceptual metrics. \citet{wang2018esrgan} has shown similar results in their SR and HR perceptual comparison.

\setlength{\tabcolsep}{4pt}
\begin{table}
\footnotesize
\begin{center}
\caption{Segmentation accuracy on the \textbf{test} set}
\label{table:segmentation}
\begin{tabular}{lllll}
\hline
\noalign{\smallskip}
  & Bicubic & SRResNet & mDCSRN $b8u4$ & mDCSRN-GAN $b8u4$\\
\noalign{\smallskip}  
\hline
\textbf{DSC} & $0.810\pm0.107$ & $0.946\pm0.047$ & \textit{$0.949\pm0.045$} & $0.929\pm0.056$ \\
\textbf{JACC} & $0.691\pm0.127$ & $0.899\pm0.061$ & \textit{$0.905\pm0.059$} & $0.871\pm0.071$ \\
\hline
\end{tabular}
\end{center}
\end{table}

\begin{figure*}
\centering
\includegraphics[width=\hsize]{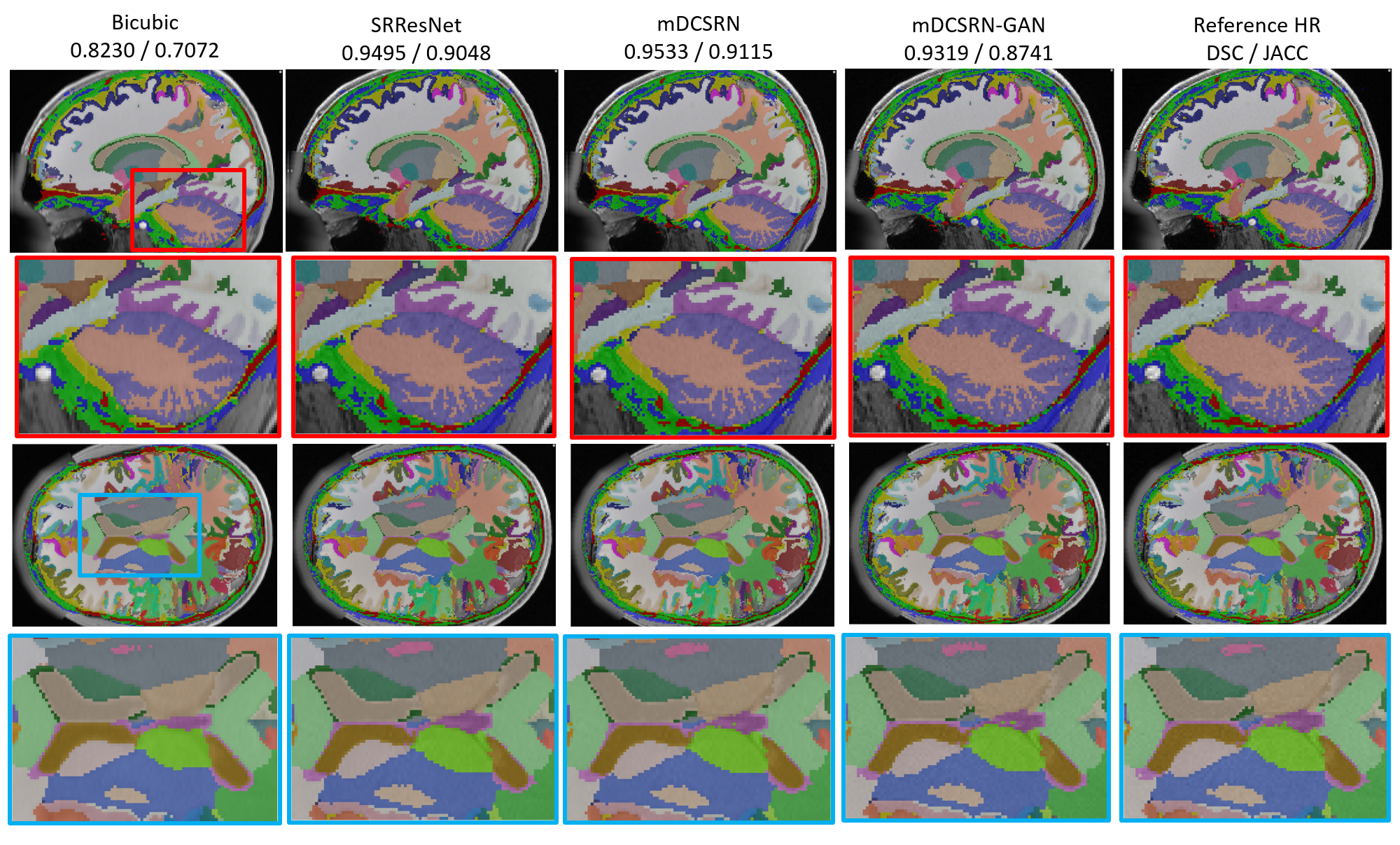}
\caption{An sample test case of segmentation from HighRes3DNet~\citep{li2017compactness} on the output of bicubic interpolation, SRResiNet, mDCSRN $b8u4$, mDCSRN-GAN $b8u4$, and Original Resolution. Average similarity metrics of this subject among 159 structures are shown on the top.}
\label{fig:segmenation}
\end{figure*}

\textbf{Segmentation Task.} We investigated the segmentation results on the output of interpolation, SRResiNet, mDCSRN, and mDCSRN-GAN. As shown in the \textbf{Table~\ref{table:segmentation}}, the segmentation results are more aligned with similarity metrics. That's because the segmentation task is more focused on the contrast instead of realistic patterns. Overall, segmentation from the SR models' output is more consistent with the segmentation of the original resolution. The high overlapping between those two indicates that segmentation on SR images are not be greatly different from those on HR. An example is shown in \textbf{Fig.~\ref{fig:segmenation}}.

\begin{figure*}
\centering
\includegraphics[width=1.0\hsize]{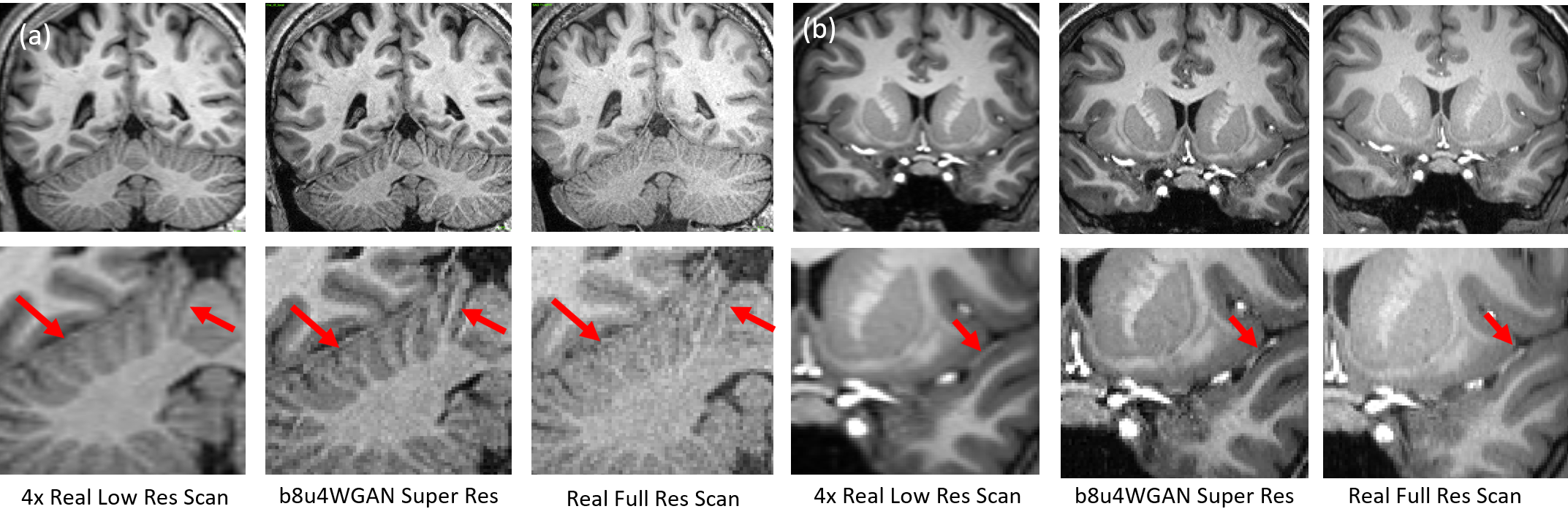}
\caption{Two \textbf{real-world} examples are shown in $2\times2$ resolution-reduced plane. There are slight mismatches between LR and HR, because they are from two separate scans. These scans were done on a different version of Siemens MRI scanner at Cedars-Sinai Medical Center. mDCSRN-GAN provides a comparable image quality to high-resolution scan.}
\label{fig:real-scan}
\end{figure*}

\begin{table}
\footnotesize
\begin{center}
\caption{Perceptual image quality metrics in \textbf{real-world} scans (N=7) }
\label{table:real-scan}
\begin{tabular}{llll}
\hline
  & NIQE\dag & MA's Score\ddag & PI\dag\\
\hline
\noalign{\smallskip}
\textbf{Low-resolution}   & $7.44\pm0.52$ & $3.83\pm0.19$ & $6.80\pm0.18$ \\
\textbf{mDCSRN-GAN}       & $5.66\pm0.77$ & $4.92\pm0.05$  & $5.37\pm0.39$ \\
\textbf{Full-resolution} & $5.17\pm0.16$ & $4.97\pm0.03$  & $5.10\pm0.07$ \\
\hline
\end{tabular}
\end{center}
{\ddag: The higher the better, ~\dag: The lower the better \par}
\end{table}

\textbf{Prospective MR Scans.} Additionally, we also performed a real-world test on seven volunteers in our on-site 3T Siemens Verio MRI scanner, which is different to those Prisma scanners utilized in the HCP dataset. We followed the same protocol as in \citet{van2013wu} except for reducing the phase encoding and slice resolution by half, which effectively reduced spatial resolution by 4x. As shown in \textbf{Fig.~\ref{fig:real-scan}} and \textbf{Table~\ref{table:real-scan}}, the mDCSRN-GAN model showed excellent ability in recovering edge details that hardly seen in the fast low-resolution scan. Besides noticeable sharpness improvement, the SR output seems to have a lower noise level and cleaner image than the original full-resolution scan because of the low-resolution image that has a better SNR than HR. It's an extra gain from super-resolution techniques in addtion to the time- and cost-saving. As the real scan was performed on a completely different machine on a different site and subject, the noise pattern and image quality were considerably different than the training dataset. It displays our model's robustness and performance in a real-world scenario. 

\section{Conclusions}

In this paper, we developed and evaluated a highly efficient architecture mDCSRN for 3D MRI SISR. We showed that the proposed mDCSRN could outperform common existing methods in voxel-based similarity matrics and segmentation accuracy with a smaller model size. We also demonstrated that with GAN-guided training, our mDCSRN-GAN could successfully recover fine details and further improve perceptual quality. Testing on prospectively acquired data showed that our model is capable of real-world clinical application. In summary, the new technique would allow a 4-fold reduction in scan time with minimal loss in image details and perceptual quality, which would substantially improve the clinical practicality of high-resolution MRI.

\clearpage

\bibliography{egbib}

\begin{thebibliography}{57}
\expandafter\ifx\csname natexlab\endcsname\relax\def\natexlab#1{#1}\fi
\providecommand{\url}[1]{\texttt{#1}}
\providecommand{\href}[2]{#2}
\providecommand{\path}[1]{#1}
\providecommand{\DOIprefix}{doi:}
\providecommand{\ArXivprefix}{arXiv:}
\providecommand{\URLprefix}{URL: }
\providecommand{\Pubmedprefix}{pmid:}
\providecommand{\doi}[1]{\href{http://dx.doi.org/#1}{\path{#1}}}
\providecommand{\Pubmed}[1]{\href{pmid:#1}{\path{#1}}}
\providecommand{\bibinfo}[2]{#2}
\ifx\xfnm\relax \def\xfnm[#1]{\unskip,\space#1}\fi
\bibitem[{Abadi et~al.(2016)Abadi, Barham, Chen, Chen, Davis, Dean, Devin,
  Ghemawat, Irving, Isard et~al.}]{abadi2016tensorflow}
\bibinfo{author}{Abadi, M.}, \bibinfo{author}{Barham, P.},
  \bibinfo{author}{Chen, J.}, \bibinfo{author}{Chen, Z.},
  \bibinfo{author}{Davis, A.}, \bibinfo{author}{Dean, J.},
  \bibinfo{author}{Devin, M.}, \bibinfo{author}{Ghemawat, S.},
  \bibinfo{author}{Irving, G.}, \bibinfo{author}{Isard, M.}, et~al.,
  \bibinfo{year}{2016}.
\newblock \bibinfo{title}{{TensorFlow}: A system for large-scale machine
  learning.}, in: \bibinfo{booktitle}{OSDI}, pp. \bibinfo{pages}{265--283}.
\bibitem[{Arjovsky et~al.(2017)Arjovsky, Chintala and
  Bottou}]{arjovsky2017wasserstein}
\bibinfo{author}{Arjovsky, M.}, \bibinfo{author}{Chintala, S.},
  \bibinfo{author}{Bottou, L.}, \bibinfo{year}{2017}.
\newblock \bibinfo{title}{Wasserstein generative adversarial networks}, in:
  \bibinfo{booktitle}{International Conference on Machine Learning}, pp.
  \bibinfo{pages}{214--223}.
\bibitem[{Avants et~al.(2009)Avants, Tustison and Song}]{avants2009advanced}
\bibinfo{author}{Avants, B.B.}, \bibinfo{author}{Tustison, N.},
  \bibinfo{author}{Song, G.}, \bibinfo{year}{2009}.
\newblock \bibinfo{title}{Advanced normalization tools ({ANTS})}.
\newblock \bibinfo{journal}{Insight j} \bibinfo{volume}{2},
  \bibinfo{pages}{1--35}.
\bibitem[{Blau et~al.(2018)Blau, Mechrez, Timofte, Michaeli and
  Zelnik-Manor}]{blau20182018}
\bibinfo{author}{Blau, Y.}, \bibinfo{author}{Mechrez, R.},
  \bibinfo{author}{Timofte, R.}, \bibinfo{author}{Michaeli, T.},
  \bibinfo{author}{Zelnik-Manor, L.}, \bibinfo{year}{2018}.
\newblock \bibinfo{title}{The 2018 {PIRM} challenge on perceptual image
  super-resolution}, in: \bibinfo{booktitle}{Proceedings of the European
  Conference on Computer Vision (ECCV)}, pp. \bibinfo{pages}{0--0}.
\bibitem[{Chen et~al.(2018)Chen, Xie, Zhou, Shi, Christodoulou and
  Li}]{chen2018brain}
\bibinfo{author}{Chen, Y.}, \bibinfo{author}{Xie, Y.}, \bibinfo{author}{Zhou,
  Z.}, \bibinfo{author}{Shi, F.}, \bibinfo{author}{Christodoulou, A.G.},
  \bibinfo{author}{Li, D.}, \bibinfo{year}{2018}.
\newblock \bibinfo{title}{Brain {MRI} super resolution using {3D} deep densely
  connected neural networks}, in: \bibinfo{booktitle}{2018 IEEE 15th
  International Symposium on Biomedical Imaging (ISBI 2018)},
  \bibinfo{organization}{IEEE}. pp. \bibinfo{pages}{739--742}.
\bibitem[{Clevert et~al.(2015)Clevert, Unterthiner and
  Hochreiter}]{clevert2015fast}
\bibinfo{author}{Clevert, D.A.}, \bibinfo{author}{Unterthiner, T.},
  \bibinfo{author}{Hochreiter, S.}, \bibinfo{year}{2015}.
\newblock \bibinfo{title}{Fast and accurate deep network learning by
  exponential linear units (elus)}.
\newblock \bibinfo{journal}{arXiv preprint arXiv:1511.07289} .
\bibitem[{Dong et~al.(2016a)Dong, Loy, He and Tang}]{dong2016image}
\bibinfo{author}{Dong, C.}, \bibinfo{author}{Loy, C.C.}, \bibinfo{author}{He,
  K.}, \bibinfo{author}{Tang, X.}, \bibinfo{year}{2016}a.
\newblock \bibinfo{title}{Image super-resolution using deep convolutional
  networks}.
\newblock \bibinfo{journal}{IEEE transactions on pattern analysis and machine
  intelligence} \bibinfo{volume}{38}, \bibinfo{pages}{295--307}.
\bibitem[{Dong et~al.(2016b)Dong, Loy and Tang}]{dong2016accelerating}
\bibinfo{author}{Dong, C.}, \bibinfo{author}{Loy, C.C.}, \bibinfo{author}{Tang,
  X.}, \bibinfo{year}{2016}b.
\newblock \bibinfo{title}{Accelerating the super-resolution convolutional
  neural network}, in: \bibinfo{booktitle}{European Conference on Computer
  Vision}, \bibinfo{organization}{Springer}. pp. \bibinfo{pages}{391--407}.
\bibitem[{Gibson et~al.(2018)Gibson, Li, Sudre, Fidon, Shakir, Wang,
  Eaton-Rosen, Gray, Doel, Hu et~al.}]{gibson2018niftynet}
\bibinfo{author}{Gibson, E.}, \bibinfo{author}{Li, W.}, \bibinfo{author}{Sudre,
  C.}, \bibinfo{author}{Fidon, L.}, \bibinfo{author}{Shakir, D.I.},
  \bibinfo{author}{Wang, G.}, \bibinfo{author}{Eaton-Rosen, Z.},
  \bibinfo{author}{Gray, R.}, \bibinfo{author}{Doel, T.}, \bibinfo{author}{Hu,
  Y.}, et~al., \bibinfo{year}{2018}.
\newblock \bibinfo{title}{{NiftyNet}: a deep-learning platform for medical
  imaging}.
\newblock \bibinfo{journal}{Computer methods and programs in biomedicine}
  \bibinfo{volume}{158}, \bibinfo{pages}{113--122}.
\bibitem[{Glasner et~al.(2009)Glasner, Bagon and Irani}]{glasner2009super}
\bibinfo{author}{Glasner, D.}, \bibinfo{author}{Bagon, S.},
  \bibinfo{author}{Irani, M.}, \bibinfo{year}{2009}.
\newblock \bibinfo{title}{Super-resolution from a single image}, in:
  \bibinfo{booktitle}{Computer Vision, 2009 IEEE 12th International Conference
  on}, \bibinfo{organization}{IEEE}. pp. \bibinfo{pages}{349--356}.
\bibitem[{Glorot and Bengio(2010)}]{glorot2010understanding}
\bibinfo{author}{Glorot, X.}, \bibinfo{author}{Bengio, Y.},
  \bibinfo{year}{2010}.
\newblock \bibinfo{title}{Understanding the difficulty of training deep
  feedforward neural networks}, in: \bibinfo{booktitle}{Proceedings of the
  Thirteenth International Conference on Artificial Intelligence and
  Statistics}, pp. \bibinfo{pages}{249--256}.
\bibitem[{Goodfellow et~al.(2014)Goodfellow, Pouget-Abadie, Mirza, Xu,
  Warde-Farley, Ozair, Courville and Bengio}]{goodfellow2014generative}
\bibinfo{author}{Goodfellow, I.}, \bibinfo{author}{Pouget-Abadie, J.},
  \bibinfo{author}{Mirza, M.}, \bibinfo{author}{Xu, B.},
  \bibinfo{author}{Warde-Farley, D.}, \bibinfo{author}{Ozair, S.},
  \bibinfo{author}{Courville, A.}, \bibinfo{author}{Bengio, Y.},
  \bibinfo{year}{2014}.
\newblock \bibinfo{title}{Generative adversarial nets}, in:
  \bibinfo{booktitle}{Advances in neural information processing systems}, pp.
  \bibinfo{pages}{2672--2680}.
\bibitem[{Greenspan(2008)}]{greenspan2008super}
\bibinfo{author}{Greenspan, H.}, \bibinfo{year}{2008}.
\newblock \bibinfo{title}{Super-resolution in medical imaging}.
\newblock \bibinfo{journal}{The Computer Journal} \bibinfo{volume}{52},
  \bibinfo{pages}{43--63}.
\bibitem[{Gulrajani et~al.(2017)Gulrajani, Ahmed, Arjovsky, Dumoulin and
  Courville}]{gulrajani2017improved}
\bibinfo{author}{Gulrajani, I.}, \bibinfo{author}{Ahmed, F.},
  \bibinfo{author}{Arjovsky, M.}, \bibinfo{author}{Dumoulin, V.},
  \bibinfo{author}{Courville, A.C.}, \bibinfo{year}{2017}.
\newblock \bibinfo{title}{Improved training of wasserstein gans}, in:
  \bibinfo{booktitle}{Advances in Neural Information Processing Systems}, pp.
  \bibinfo{pages}{5769--5779}.
\bibitem[{He et~al.(2016)He, Zhang, Ren and Sun}]{he2016deep}
\bibinfo{author}{He, K.}, \bibinfo{author}{Zhang, X.}, \bibinfo{author}{Ren,
  S.}, \bibinfo{author}{Sun, J.}, \bibinfo{year}{2016}.
\newblock \bibinfo{title}{Deep residual learning for image recognition}, in:
  \bibinfo{booktitle}{Proceedings of the IEEE conference on computer vision and
  pattern recognition}, pp. \bibinfo{pages}{770--778}.
\bibitem[{Huang et~al.(2017)Huang, Liu, Weinberger and van~der
  Maaten}]{huang2017densely}
\bibinfo{author}{Huang, G.}, \bibinfo{author}{Liu, Z.},
  \bibinfo{author}{Weinberger, K.Q.}, \bibinfo{author}{van~der Maaten, L.},
  \bibinfo{year}{2017}.
\newblock \bibinfo{title}{Densely connected convolutional networks}, in:
  \bibinfo{booktitle}{Proceedings of the IEEE conference on computer vision and
  pattern recognition}, p.~\bibinfo{pages}{3}.
\bibitem[{Hubel and Wiesel(1962)}]{hubel1962receptive}
\bibinfo{author}{Hubel, D.H.}, \bibinfo{author}{Wiesel, T.N.},
  \bibinfo{year}{1962}.
\newblock \bibinfo{title}{Receptive fields, binocular interaction and
  functional architecture in the cat's visual cortex}.
\newblock \bibinfo{journal}{The Journal of physiology} \bibinfo{volume}{160},
  \bibinfo{pages}{106--154}.
\bibitem[{Ioffe and Szegedy(2015)}]{ioffe2015batch}
\bibinfo{author}{Ioffe, S.}, \bibinfo{author}{Szegedy, C.},
  \bibinfo{year}{2015}.
\newblock \bibinfo{title}{Batch normalization: Accelerating deep network
  training by reducing internal covariate shift}, in:
  \bibinfo{booktitle}{International conference on machine learning}, pp.
  \bibinfo{pages}{448--456}.
\bibitem[{Jaccard(1901)}]{jaccard1901etude}
\bibinfo{author}{Jaccard, P.}, \bibinfo{year}{1901}.
\newblock \bibinfo{title}{{\'E}tude comparative de la distribution florale dans
  une portion des alpes et des jura}.
\newblock \bibinfo{journal}{Bull Soc Vaudoise Sci Nat} \bibinfo{volume}{37},
  \bibinfo{pages}{547--579}.
\bibitem[{Johnson et~al.(2016)Johnson, Alahi and
  Fei-Fei}]{johnson2016perceptual}
\bibinfo{author}{Johnson, J.}, \bibinfo{author}{Alahi, A.},
  \bibinfo{author}{Fei-Fei, L.}, \bibinfo{year}{2016}.
\newblock \bibinfo{title}{Perceptual losses for real-time style transfer and
  super-resolution}, in: \bibinfo{booktitle}{European Conference on Computer
  Vision}, \bibinfo{organization}{Springer}. pp. \bibinfo{pages}{694--711}.
\bibitem[{Kim et~al.(2016)Kim, Kwon~Lee and Mu~Lee}]{kim2016accurate}
\bibinfo{author}{Kim, J.}, \bibinfo{author}{Kwon~Lee, J.},
  \bibinfo{author}{Mu~Lee, K.}, \bibinfo{year}{2016}.
\newblock \bibinfo{title}{Accurate image super-resolution using very deep
  convolutional networks}, in: \bibinfo{booktitle}{Proceedings of the IEEE
  Conference on Computer Vision and Pattern Recognition}, pp.
  \bibinfo{pages}{1646--1654}.
\bibitem[{Kingma and Ba()}]{kingma1412adam}
\bibinfo{author}{Kingma, D.P.}, \bibinfo{author}{Ba, J.}, .
\newblock \bibinfo{title}{Adam: A method for stochastic optimization}, in:
  \bibinfo{booktitle}{Proceedings of the 3rd International Conference on
  Learning Representations (ICLR), arXiv preprint arXiv}.
\bibitem[{Ledig et~al.(2017)Ledig, Theis, Husz{\'a}r, Caballero, Cunningham,
  Acosta, Aitken, Tejani, Totz, Wang et~al.}]{ledig2017photo}
\bibinfo{author}{Ledig, C.}, \bibinfo{author}{Theis, L.},
  \bibinfo{author}{Husz{\'a}r, F.}, \bibinfo{author}{Caballero, J.},
  \bibinfo{author}{Cunningham, A.}, \bibinfo{author}{Acosta, A.},
  \bibinfo{author}{Aitken, A.}, \bibinfo{author}{Tejani, A.},
  \bibinfo{author}{Totz, J.}, \bibinfo{author}{Wang, Z.}, et~al.,
  \bibinfo{year}{2017}.
\newblock \bibinfo{title}{Photo-realistic single image super-resolution using a
  generative adversarial network}, in: \bibinfo{booktitle}{Proceedings of the
  IEEE conference on computer vision and pattern recognition}, pp.
  \bibinfo{pages}{4681--4690}.
\bibitem[{Li et~al.(2017)Li, Wang, Fidon, Ourselin, Cardoso and
  Vercauteren}]{li2017compactness}
\bibinfo{author}{Li, W.}, \bibinfo{author}{Wang, G.}, \bibinfo{author}{Fidon,
  L.}, \bibinfo{author}{Ourselin, S.}, \bibinfo{author}{Cardoso, M.J.},
  \bibinfo{author}{Vercauteren, T.}, \bibinfo{year}{2017}.
\newblock \bibinfo{title}{On the compactness, efficiency, and representation of
  {3D} convolutional networks: brain parcellation as a pretext task}, in:
  \bibinfo{booktitle}{International Conference on Information Processing in
  Medical Imaging}, \bibinfo{organization}{Springer}. pp.
  \bibinfo{pages}{348--360}.
\bibitem[{Lim et~al.(2017)Lim, Son, Kim, Nah and Lee}]{lim2017enhanced}
\bibinfo{author}{Lim, B.}, \bibinfo{author}{Son, S.}, \bibinfo{author}{Kim,
  H.}, \bibinfo{author}{Nah, S.}, \bibinfo{author}{Lee, K.M.},
  \bibinfo{year}{2017}.
\newblock \bibinfo{title}{Enhanced deep residual networks for single image
  super-resolution}, in: \bibinfo{booktitle}{The IEEE Conference on Computer
  Vision and Pattern Recognition (CVPR) Workshops}, p.~\bibinfo{pages}{3}.
\bibitem[{Litjens et~al.(2017)Litjens, Kooi, Bejnordi, Setio, Ciompi,
  Ghafoorian, van~der Laak, van Ginneken and S{\'a}nchez}]{litjens2017survey}
\bibinfo{author}{Litjens, G.}, \bibinfo{author}{Kooi, T.},
  \bibinfo{author}{Bejnordi, B.E.}, \bibinfo{author}{Setio, A.A.A.},
  \bibinfo{author}{Ciompi, F.}, \bibinfo{author}{Ghafoorian, M.},
  \bibinfo{author}{van~der Laak, J.A.}, \bibinfo{author}{van Ginneken, B.},
  \bibinfo{author}{S{\'a}nchez, C.I.}, \bibinfo{year}{2017}.
\newblock \bibinfo{title}{A survey on deep learning in medical image analysis}.
\newblock \bibinfo{journal}{Medical image analysis} \bibinfo{volume}{42},
  \bibinfo{pages}{60--88}.
\bibitem[{Liu et~al.(2016)Liu, Wang, Nasrabadi and Huang}]{liu2016learning}
\bibinfo{author}{Liu, D.}, \bibinfo{author}{Wang, Z.},
  \bibinfo{author}{Nasrabadi, N.}, \bibinfo{author}{Huang, T.},
  \bibinfo{year}{2016}.
\newblock \bibinfo{title}{Learning a mixture of deep networks for single image
  super-resolution}, in: \bibinfo{booktitle}{Asian Conference on Computer
  Vision}, \bibinfo{organization}{Springer}. pp. \bibinfo{pages}{145--156}.
\bibitem[{Ma et~al.(2017)Ma, Yang, Yang and Yang}]{ma2017learning}
\bibinfo{author}{Ma, C.}, \bibinfo{author}{Yang, C.Y.}, \bibinfo{author}{Yang,
  X.}, \bibinfo{author}{Yang, M.H.}, \bibinfo{year}{2017}.
\newblock \bibinfo{title}{Learning a no-reference quality metric for
  single-image super-resolution}.
\newblock \bibinfo{journal}{Computer Vision and Image Understanding}
  \bibinfo{volume}{158}, \bibinfo{pages}{1--16}.
\bibitem[{Mittal et~al.(2012)Mittal, Soundararajan and
  Bovik}]{mittal2012making}
\bibinfo{author}{Mittal, A.}, \bibinfo{author}{Soundararajan, R.},
  \bibinfo{author}{Bovik, A.C.}, \bibinfo{year}{2012}.
\newblock \bibinfo{title}{Making a “completely blind” image quality
  analyzer}.
\newblock \bibinfo{journal}{IEEE Signal Processing Letters}
  \bibinfo{volume}{20}, \bibinfo{pages}{209--212}.
\bibitem[{Oktay et~al.(2016)Oktay, Bai, Lee, Guerrero, Kamnitsas, Caballero,
  de~Marvao, Cook, O’Regan and Rueckert}]{oktay2016multi}
\bibinfo{author}{Oktay, O.}, \bibinfo{author}{Bai, W.}, \bibinfo{author}{Lee,
  M.}, \bibinfo{author}{Guerrero, R.}, \bibinfo{author}{Kamnitsas, K.},
  \bibinfo{author}{Caballero, J.}, \bibinfo{author}{de~Marvao, A.},
  \bibinfo{author}{Cook, S.}, \bibinfo{author}{O’Regan, D.},
  \bibinfo{author}{Rueckert, D.}, \bibinfo{year}{2016}.
\newblock \bibinfo{title}{Multi-input cardiac image super-resolution using
  convolutional neural networks}, in: \bibinfo{booktitle}{International
  Conference on Medical Image Computing and Computer-Assisted Intervention},
  \bibinfo{organization}{Springer}. pp. \bibinfo{pages}{246--254}.
\bibitem[{Pang et~al.(2016)Pang, Chen, Fan, Nguyen, Yang, Xie and
  Li}]{pang2016high}
\bibinfo{author}{Pang, J.}, \bibinfo{author}{Chen, Y.}, \bibinfo{author}{Fan,
  Z.}, \bibinfo{author}{Nguyen, C.}, \bibinfo{author}{Yang, Q.},
  \bibinfo{author}{Xie, Y.}, \bibinfo{author}{Li, D.}, \bibinfo{year}{2016}.
\newblock \bibinfo{title}{High efficiency coronary {MR} angiography with
  nonrigid cardiac motion correction}.
\newblock \bibinfo{journal}{Magnetic Resonance in Medicine}
  \bibinfo{volume}{76}, \bibinfo{pages}{1345--1353}.
\bibitem[{Park et~al.(2003)Park, Park and Kang}]{park2003super}
\bibinfo{author}{Park, S.C.}, \bibinfo{author}{Park, M.K.},
  \bibinfo{author}{Kang, M.G.}, \bibinfo{year}{2003}.
\newblock \bibinfo{title}{Super-resolution image reconstruction: a technical
  overview}.
\newblock \bibinfo{journal}{IEEE signal processing magazine}
  \bibinfo{volume}{20}, \bibinfo{pages}{21--36}.
\bibitem[{Pham et~al.(2017)Pham, Ducournau, Fablet and
  Rousseau}]{pham2017brain}
\bibinfo{author}{Pham, C.H.}, \bibinfo{author}{Ducournau, A.},
  \bibinfo{author}{Fablet, R.}, \bibinfo{author}{Rousseau, F.},
  \bibinfo{year}{2017}.
\newblock \bibinfo{title}{Brain {MRI} super-resolution using deep {3D}
  convolutional networks}, in: \bibinfo{booktitle}{Biomedical Imaging (ISBI
  2017), 2017 IEEE 14th International Symposium on},
  \bibinfo{organization}{IEEE}. pp. \bibinfo{pages}{197--200}.
\bibitem[{Pruessner et~al.(2000)Pruessner, Li, Serles, Pruessner, Collins,
  Kabani, Lupien and Evans}]{pruessner2000volumetry}
\bibinfo{author}{Pruessner, J.C.}, \bibinfo{author}{Li, L.M.},
  \bibinfo{author}{Serles, W.}, \bibinfo{author}{Pruessner, M.},
  \bibinfo{author}{Collins, D.L.}, \bibinfo{author}{Kabani, N.},
  \bibinfo{author}{Lupien, S.}, \bibinfo{author}{Evans, A.C.},
  \bibinfo{year}{2000}.
\newblock \bibinfo{title}{Volumetry of hippocampus and amygdala with
  high-resolution {MRI} and three-dimensional analysis software: minimizing the
  discrepancies between laboratories}.
\newblock \bibinfo{journal}{Cerebral Cortex} \bibinfo{volume}{10},
  \bibinfo{pages}{433--442}.
\bibitem[{Rudin et~al.(1992)Rudin, Osher and Fatemi}]{rudin1992nonlinear}
\bibinfo{author}{Rudin, L.I.}, \bibinfo{author}{Osher, S.},
  \bibinfo{author}{Fatemi, E.}, \bibinfo{year}{1992}.
\newblock \bibinfo{title}{Nonlinear total variation based noise removal
  algorithms}.
\newblock \bibinfo{journal}{Physica D: nonlinear phenomena}
  \bibinfo{volume}{60}, \bibinfo{pages}{259--268}.
\bibitem[{Salimans et~al.(2016)Salimans, Goodfellow, Zaremba, Cheung, Radford
  and Chen}]{salimans2016improved}
\bibinfo{author}{Salimans, T.}, \bibinfo{author}{Goodfellow, I.},
  \bibinfo{author}{Zaremba, W.}, \bibinfo{author}{Cheung, V.},
  \bibinfo{author}{Radford, A.}, \bibinfo{author}{Chen, X.},
  \bibinfo{year}{2016}.
\newblock \bibinfo{title}{Improved techniques for training gans}, in:
  \bibinfo{booktitle}{Advances in Neural Information Processing Systems}, pp.
  \bibinfo{pages}{2234--2242}.
\bibitem[{Shi et~al.(2015)Shi, Cheng, Wang, Yap and Shen}]{shi2015lrtv}
\bibinfo{author}{Shi, F.}, \bibinfo{author}{Cheng, J.}, \bibinfo{author}{Wang,
  L.}, \bibinfo{author}{Yap, P.T.}, \bibinfo{author}{Shen, D.},
  \bibinfo{year}{2015}.
\newblock \bibinfo{title}{{LRTV}: {MR} image super-resolution with low-rank and
  total variation regularizations}.
\newblock \bibinfo{journal}{IEEE transactions on medical imaging}
  \bibinfo{volume}{34}, \bibinfo{pages}{2459--2466}.
\bibitem[{S{\o}rensen(1948)}]{sorensen1948method}
\bibinfo{author}{S{\o}rensen, T.}, \bibinfo{year}{1948}.
\newblock \bibinfo{title}{A method of establishing groups of equal amplitude in
  plant sociology based on similarity of species content and its application to
  analyses of the vegetation on danish commons}.
\newblock \bibinfo{journal}{Biologiske Skrifter} \bibinfo{volume}{5},
  \bibinfo{pages}{1--34}.
\bibitem[{Srivastava et~al.(2015)Srivastava, Greff and
  Schmidhuber}]{srivastava2015training}
\bibinfo{author}{Srivastava, R.K.}, \bibinfo{author}{Greff, K.},
  \bibinfo{author}{Schmidhuber, J.}, \bibinfo{year}{2015}.
\newblock \bibinfo{title}{Training very deep networks}, in:
  \bibinfo{booktitle}{Advances in neural information processing systems}, pp.
  \bibinfo{pages}{2377--2385}.
\bibitem[{Stucht et~al.(2015)Stucht, Danishad, Schulze, Godenschweger, Zaitsev
  and Speck}]{stucht2015highest}
\bibinfo{author}{Stucht, D.}, \bibinfo{author}{Danishad, K.A.},
  \bibinfo{author}{Schulze, P.}, \bibinfo{author}{Godenschweger, F.},
  \bibinfo{author}{Zaitsev, M.}, \bibinfo{author}{Speck, O.},
  \bibinfo{year}{2015}.
\newblock \bibinfo{title}{Highest resolution in vivo human brain {MRI} using
  prospective motion correction}.
\newblock \bibinfo{journal}{PloS one} \bibinfo{volume}{10},
  \bibinfo{pages}{e0133921}.
\bibitem[{Sun et~al.(2016)Sun, Chen, Wang, Liu and Liu}]{sun2016depth}
\bibinfo{author}{Sun, S.}, \bibinfo{author}{Chen, W.}, \bibinfo{author}{Wang,
  L.}, \bibinfo{author}{Liu, X.}, \bibinfo{author}{Liu, T.Y.},
  \bibinfo{year}{2016}.
\newblock \bibinfo{title}{On the depth of deep neural networks: A theoretical
  view.}, in: \bibinfo{booktitle}{AAAI}, pp. \bibinfo{pages}{2066--2072}.
\bibitem[{Szegedy et~al.(2017)Szegedy, Ioffe, Vanhoucke and
  Alemi}]{szegedy2017inception}
\bibinfo{author}{Szegedy, C.}, \bibinfo{author}{Ioffe, S.},
  \bibinfo{author}{Vanhoucke, V.}, \bibinfo{author}{Alemi, A.A.},
  \bibinfo{year}{2017}.
\newblock \bibinfo{title}{{Inception-v4}, inception-resnet and the impact of
  residual connections on learning.}, in: \bibinfo{booktitle}{AAAI},
  p.~\bibinfo{pages}{12}.
\bibitem[{Szegedy et~al.(2016)Szegedy, Vanhoucke, Ioffe, Shlens and
  Wojna}]{szegedy2016rethinking}
\bibinfo{author}{Szegedy, C.}, \bibinfo{author}{Vanhoucke, V.},
  \bibinfo{author}{Ioffe, S.}, \bibinfo{author}{Shlens, J.},
  \bibinfo{author}{Wojna, Z.}, \bibinfo{year}{2016}.
\newblock \bibinfo{title}{Rethinking the inception architecture for computer
  vision}, in: \bibinfo{booktitle}{Proceedings of the IEEE Conference on
  Computer Vision and Pattern Recognition}, pp. \bibinfo{pages}{2818--2826}.
\bibitem[{Tai et~al.(2017)Tai, Yang and Liu}]{tai2017image}
\bibinfo{author}{Tai, Y.}, \bibinfo{author}{Yang, J.}, \bibinfo{author}{Liu,
  X.}, \bibinfo{year}{2017}.
\newblock \bibinfo{title}{Image super-resolution via deep recursive residual
  network}, in: \bibinfo{booktitle}{The IEEE Conference on Computer Vision and
  Pattern Recognition (CVPR)}.
\bibitem[{Tong et~al.(2017)Tong, Li, Liu and Gao}]{tong2017image}
\bibinfo{author}{Tong, T.}, \bibinfo{author}{Li, G.}, \bibinfo{author}{Liu,
  X.}, \bibinfo{author}{Gao, Q.}, \bibinfo{year}{2017}.
\newblock \bibinfo{title}{Image super-resolution using dense skip connections},
  in: \bibinfo{booktitle}{2017 IEEE International Conference on Computer Vision
  (ICCV)}, \bibinfo{organization}{IEEE}. pp. \bibinfo{pages}{4809--4817}.
\bibitem[{Tustison et~al.(2010)Tustison, Avants, Cook, Zheng, Egan, Yushkevich
  and Gee}]{tustison2010n4itk}
\bibinfo{author}{Tustison, N.J.}, \bibinfo{author}{Avants, B.B.},
  \bibinfo{author}{Cook, P.A.}, \bibinfo{author}{Zheng, Y.},
  \bibinfo{author}{Egan, A.}, \bibinfo{author}{Yushkevich, P.A.},
  \bibinfo{author}{Gee, J.C.}, \bibinfo{year}{2010}.
\newblock \bibinfo{title}{{N4ITK}: improved {N3} bias correction}.
\newblock \bibinfo{journal}{IEEE transactions on medical imaging}
  \bibinfo{volume}{29}, \bibinfo{pages}{1310}.
\bibitem[{Van~Essen et~al.(2013)Van~Essen, Smith, Barch, Behrens, Yacoub,
  Ugurbil, Consortium et~al.}]{van2013wu}
\bibinfo{author}{Van~Essen, D.C.}, \bibinfo{author}{Smith, S.M.},
  \bibinfo{author}{Barch, D.M.}, \bibinfo{author}{Behrens, T.E.},
  \bibinfo{author}{Yacoub, E.}, \bibinfo{author}{Ugurbil, K.},
  \bibinfo{author}{Consortium, W.M.H.}, et~al., \bibinfo{year}{2013}.
\newblock \bibinfo{title}{The {WU-Minn} human connectome project: an overview}.
\newblock \bibinfo{journal}{Neuroimage} \bibinfo{volume}{80},
  \bibinfo{pages}{62--79}.
\bibitem[{Veit et~al.(2016)Veit, Wilber and Belongie}]{veit2016residual}
\bibinfo{author}{Veit, A.}, \bibinfo{author}{Wilber, M.J.},
  \bibinfo{author}{Belongie, S.}, \bibinfo{year}{2016}.
\newblock \bibinfo{title}{Residual networks behave like ensembles of relatively
  shallow networks}, in: \bibinfo{booktitle}{Advances in Neural Information
  Processing Systems}, pp. \bibinfo{pages}{550--558}.
\bibitem[{Venkatanath et~al.(2015)Venkatanath, Praneeth, Bh, Channappayya and
  Medasani}]{venkatanath2015blind}
\bibinfo{author}{Venkatanath, N.}, \bibinfo{author}{Praneeth, D.},
  \bibinfo{author}{Bh, M.C.}, \bibinfo{author}{Channappayya, S.S.},
  \bibinfo{author}{Medasani, S.S.}, \bibinfo{year}{2015}.
\newblock \bibinfo{title}{Blind image quality evaluation using perception based
  features}, in: \bibinfo{booktitle}{2015 Twenty First National Conference on
  Communications (NCC)}, \bibinfo{organization}{IEEE}. pp.
  \bibinfo{pages}{1--6}.
\bibitem[{Wang et~al.(2016)Wang, Su, Ying, Peng, Zhu, Liang, Feng and
  Liang}]{wang2016accelerating}
\bibinfo{author}{Wang, S.}, \bibinfo{author}{Su, Z.}, \bibinfo{author}{Ying,
  L.}, \bibinfo{author}{Peng, X.}, \bibinfo{author}{Zhu, S.},
  \bibinfo{author}{Liang, F.}, \bibinfo{author}{Feng, D.},
  \bibinfo{author}{Liang, D.}, \bibinfo{year}{2016}.
\newblock \bibinfo{title}{Accelerating magnetic resonance imaging via deep
  learning}, in: \bibinfo{booktitle}{Biomedical Imaging (ISBI), 2016 IEEE 13th
  International Symposium on}, \bibinfo{organization}{IEEE}. pp.
  \bibinfo{pages}{514--517}.
\bibitem[{Wang et~al.(2018)Wang, Yu, Wu, Gu, Liu, Dong, Qiao and
  Change~Loy}]{wang2018esrgan}
\bibinfo{author}{Wang, X.}, \bibinfo{author}{Yu, K.}, \bibinfo{author}{Wu, S.},
  \bibinfo{author}{Gu, J.}, \bibinfo{author}{Liu, Y.}, \bibinfo{author}{Dong,
  C.}, \bibinfo{author}{Qiao, Y.}, \bibinfo{author}{Change~Loy, C.},
  \bibinfo{year}{2018}.
\newblock \bibinfo{title}{{ESRGAN}: Enhanced super-resolution generative
  adversarial networks}, in: \bibinfo{booktitle}{Proceedings of the European
  Conference on Computer Vision (ECCV)}, pp. \bibinfo{pages}{0--0}.
\bibitem[{Wang et~al.(2004)Wang, Bovik, Sheikh and Simoncelli}]{wang2004image}
\bibinfo{author}{Wang, Z.}, \bibinfo{author}{Bovik, A.C.},
  \bibinfo{author}{Sheikh, H.R.}, \bibinfo{author}{Simoncelli, E.P.},
  \bibinfo{year}{2004}.
\newblock \bibinfo{title}{Image quality assessment: from error visibility to
  structural similarity}.
\newblock \bibinfo{journal}{IEEE transactions on image processing}
  \bibinfo{volume}{13}, \bibinfo{pages}{600--612}.
\bibitem[{Xie et~al.(2016)Xie, Wisse, Das, Wang, Wolk, Manj{\'o}n and
  Yushkevich}]{xie2016accounting}
\bibinfo{author}{Xie, L.}, \bibinfo{author}{Wisse, L.E.}, \bibinfo{author}{Das,
  S.R.}, \bibinfo{author}{Wang, H.}, \bibinfo{author}{Wolk, D.A.},
  \bibinfo{author}{Manj{\'o}n, J.V.}, \bibinfo{author}{Yushkevich, P.A.},
  \bibinfo{year}{2016}.
\newblock \bibinfo{title}{Accounting for the confound of meninges in segmenting
  entorhinal and perirhinal cortices in {T1}-weighted {MRI}}, in:
  \bibinfo{booktitle}{International Conference on Medical Image Computing and
  Computer-assisted Intervention}, \bibinfo{organization}{Springer}. pp.
  \bibinfo{pages}{564--571}.
\bibitem[{Yang et~al.(2014)Yang, Ma and Yang}]{yang2014single}
\bibinfo{author}{Yang, C.Y.}, \bibinfo{author}{Ma, C.}, \bibinfo{author}{Yang,
  M.H.}, \bibinfo{year}{2014}.
\newblock \bibinfo{title}{Single-image super-resolution: A benchmark}, in:
  \bibinfo{booktitle}{European Conference on Computer Vision},
  \bibinfo{organization}{Springer}. pp. \bibinfo{pages}{372--386}.
\bibitem[{Yang et~al.(2016)Yang, Sharif, Pang, Kali, Bi, Cokic, Li and
  Dharmakumar}]{yang2016free}
\bibinfo{author}{Yang, H.J.}, \bibinfo{author}{Sharif, B.},
  \bibinfo{author}{Pang, J.}, \bibinfo{author}{Kali, A.}, \bibinfo{author}{Bi,
  X.}, \bibinfo{author}{Cokic, I.}, \bibinfo{author}{Li, D.},
  \bibinfo{author}{Dharmakumar, R.}, \bibinfo{year}{2016}.
\newblock \bibinfo{title}{Free-breathing, motion-corrected, highly efficient
  whole heart {T2} mapping at {3T} with hybrid radial-cartesian trajectory}.
\newblock \bibinfo{journal}{Magnetic Resonance in Medicine}
  \bibinfo{volume}{75}, \bibinfo{pages}{126--136}.
\bibitem[{You et~al.(2019)You, Li, Zhang, Zhang, Shan, Li, Ju, Zhao, Zhang,
  Cong et~al.}]{you2019ct}
\bibinfo{author}{You, C.}, \bibinfo{author}{Li, G.}, \bibinfo{author}{Zhang,
  Y.}, \bibinfo{author}{Zhang, X.}, \bibinfo{author}{Shan, H.},
  \bibinfo{author}{Li, M.}, \bibinfo{author}{Ju, S.}, \bibinfo{author}{Zhao,
  Z.}, \bibinfo{author}{Zhang, Z.}, \bibinfo{author}{Cong, W.}, et~al.,
  \bibinfo{year}{2019}.
\newblock \bibinfo{title}{{CT} super-resolution {GAN} constrained by the
  identical, residual, and cycle learning ensemble ({GAN-CIRCLE})}.
\newblock \bibinfo{journal}{IEEE Transactions on Medical Imaging}
  \bibinfo{volume}{39}, \bibinfo{pages}{188--203}.
\bibitem[{Zhou et~al.(2017)Zhou, Nguyen, Chen, Shaw, Deng, Xie, Dawkins,
  Marb{\'a}n and Li}]{zhou2017optimized}
\bibinfo{author}{Zhou, Z.}, \bibinfo{author}{Nguyen, C.},
  \bibinfo{author}{Chen, Y.}, \bibinfo{author}{Shaw, J.L.},
  \bibinfo{author}{Deng, Z.}, \bibinfo{author}{Xie, Y.},
  \bibinfo{author}{Dawkins, J.}, \bibinfo{author}{Marb{\'a}n, E.},
  \bibinfo{author}{Li, D.}, \bibinfo{year}{2017}.
\newblock \bibinfo{title}{Optimized cest cardiovascular magnetic resonance for
  assessment of metabolic activity in the heart}.
\newblock \bibinfo{journal}{Journal of Cardiovascular Magnetic Resonance}
  \bibinfo{volume}{19}, \bibinfo{pages}{95}.

\end{thebibliography}

\end{document}